\relax
\documentclass[letterpaper]{article} % DO NOT CHANGE THIS

 \newlength\titlebox 
 \twocolumn 
\usepackage{natbib}
\let\cite\citep

\date{}

\usepackage{times}  % DO NOT CHANGE THIS
\usepackage{helvet} % DO NOT CHANGE THIS
\usepackage{courier}  % DO NOT CHANGE THIS
\usepackage[hyphens]{url}  % DO NOT CHANGE THIS
\usepackage{graphicx} % DO NOT CHANGE THIS
\usepackage{graphicx}  % DO NOT CHANGE THIS
\usepackage{natbib}  % DO NOT CHANGE THIS AND DO NOT ADD ANY OPTIONS TO IT
\usepackage{caption} % DO NOT CHANGE THIS AND DO NOT ADD ANY OPTIONS TO IT
\usepackage{booktabs}
\usepackage{amssymb}
\usepackage[ruled,vlined,linesnumbered]{algorithm2e}
\usepackage{amsmath}
\usepackage[table]{xcolor}

\usepackage{cleveref}
\usepackage{dblfloatfix} 
\usepackage{appendix}

\usepackage[ruled,vlined]{algorithm2e}
\usepackage{natbib}
\usepackage{graphicx}
\usepackage{xcolor}
\usepackage{amsmath}
\usepackage{amssymb}
\usepackage{wrapfig}
\usepackage{cleveref}
\usepackage{booktabs}
\usepackage{tabularx}
\usepackage{multirow}
\usepackage{amssymb}

\title{Understanding Decoupled and Early Weight Decay}
\author{
    Johan Bjorck,
    Kilian Weinberger,
    Carla Gomes\\
    Cornell University
    
}
\begin{document}
\maketitle
% content
\begin{abstract}
Weight decay (WD) is a traditional regularization technique in deep learning, but despite its ubiquity, its behavior is still an area of active research. Golatkar et al. have recently shown that WD only matters at the start of the training in computer vision, upending traditional wisdom. Loshchilov et al. show that for adaptive optimizers, manually decaying weights can outperform adding an $l_2$ penalty to the loss. This technique has become increasingly popular and is referred to as decoupled WD. The goal of this paper is to investigate these two recent empirical observations. We demonstrate that by applying WD only at the start, the network norm stays small throughout training. This has a regularizing effect as the effective gradient updates become larger. However, traditional generalizations metrics fail to capture this effect of WD, and we show how a simple scale-invariant metric can. We also show how the growth of network weights is heavily influenced by the dataset and its generalization properties. For decoupled WD, we perform experiments in NLP and RL where adaptive optimizers are the norm. We demonstrate that the primary issue that decoupled WD alleviates is the mixing of gradients from the objective function and the $l_2$ penalty in the buffers of Adam (which stores the estimates of the first-order moment). Adaptivity itself is not problematic and decoupled WD ensures that the gradients from the $l_2$ term cannot "drown out" the true objective, facilitating easier hyperparameter tuning.

\end{abstract}

\begin{figure*}[t!]
\centering
\includegraphics[width=0.9\textwidth]{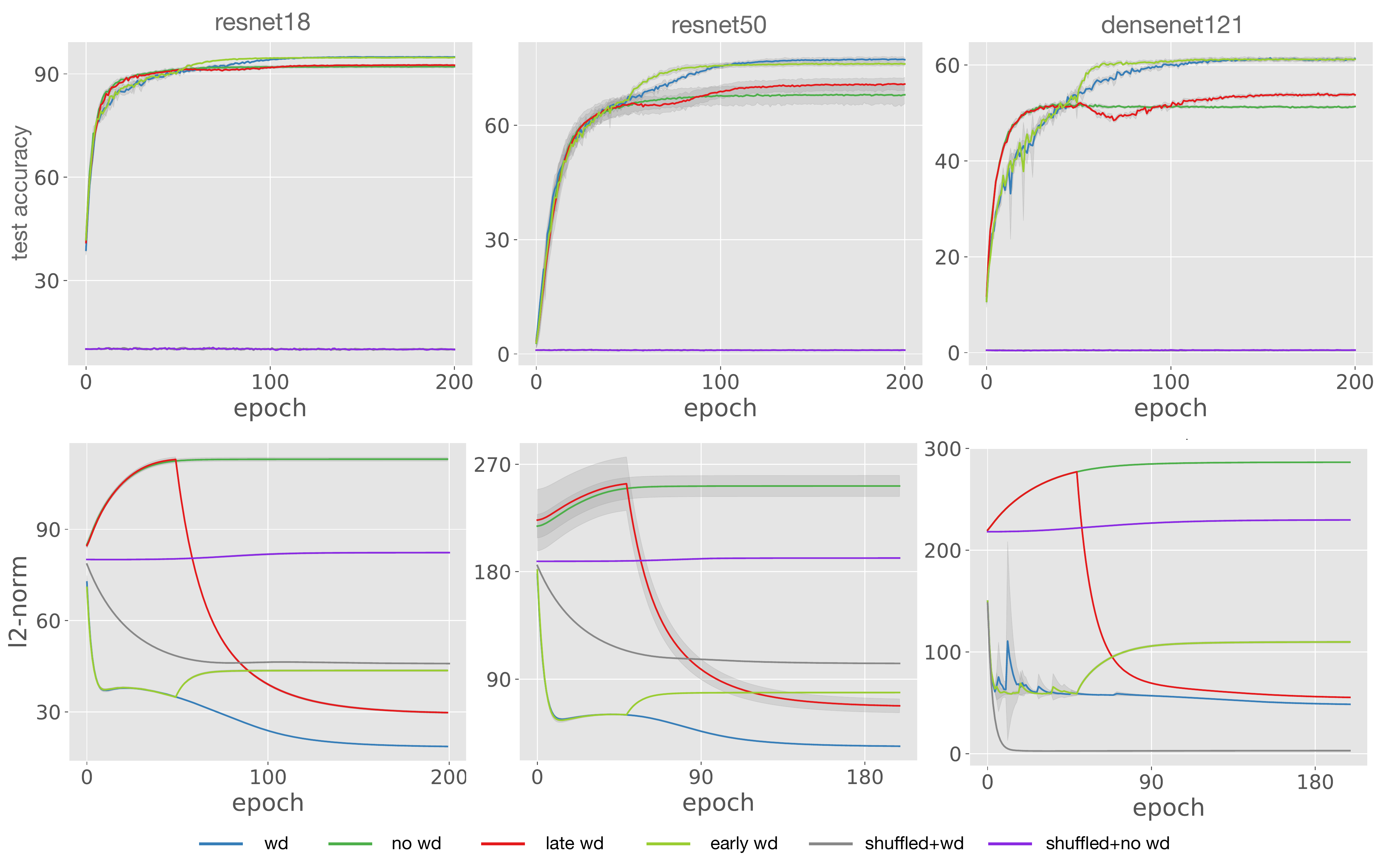}
\caption{\textit{(Top.)} \citet{golatkar2019time} have shown that for image classification, starting WD only after epoch 50 brings little benefit, whereas stopping it after epoch 50 performs on par with using it throughout the training. \textit{(Bottom.)} The $l_2$ norm of the weights increases dramatically at the start. Applying WD only during early parts of training ensures small weights throughout the optimization process. By applying WD late, it takes many epochs for the norm to shrink. We also plot curves for networks trained on datasets with shuffled labels and note that weight norms grow less under such settings.}
\label{fig:cifar_norms_over_time}
\end{figure*}

\section{Introduction}

The roots of weight decay (WD) go back to at least \citet{tikhonov1943stability}, and within the context of deep learning, it has been used at least since 1987 \cite{hinton1987learning}. Modern DNNs are typically trained with WD \cite{tan2019efficientnet, huang2017densely}. The technique is also used in modern NLP (natural language processing) \cite{ott2019fairseq, radford2018improving} but is less commonly used in reinforcement learning. Despite its ubiquity, there is still ongoing research on WD -- \citet{golatkar2019time} have recently shown that WD essentially only matters at the start of the training in computer vision. Additionally, \citet{loshchilov2017decoupled} have shown that WD interacts poorly with adaptive optimizers. The motivation of this paper is to investigate and explain these recent empirical observations on WD. It is common to formulate WD as adding an $l_2$ penalty $ \frac{1}{2}\lambda \| w \|_2^2$ to a loss function $L(w) = \frac{1}{|D|} \sum_{i \in D} \ell_i(w)$ for a dataset $D$ and weights $w$. For SGD with batch $B$ and learning rate $\alpha$ this leads to the following update
\begin{equation}
\label{eq:def_update}
w_{t+1} = w_t - \frac{\alpha}{|B|} \nabla  \sum_{i \in B} \ell_i(w_t) - \alpha \lambda w_t
\end{equation}
By adding an $l_2$ penalty term, the weights $w$ are "decayed" by a factor $(1-\alpha \lambda)$ per update. Thus, it is common to use the terms weight decay and $l_2$ regularization interchangeably.

\textbf{Background.} The motivation for this work is to understand two recent observations. The first observation comes from \citet{loshchilov2017decoupled}, who show that for Adam \cite{kingma2014adam}, manually decaying weights can outperform an $l_2$ loss. As the gradient of the $l_2$ term will appear both in the numerator and denominator of the adaptive gradient step, these methods are not equivalent. \citet{loshchilov2017decoupled} dub this technique decoupled weight decay and perform experiments on small-scale computer vision tasks, observing improved generalization and increased hyperparameter stability. This strategy has become increasingly popular \cite{wang2018sface, radford2018improving, carion2020end, liu2020crisisbert} and is e.g. used in the Facebook NLP repository fairseq \cite{ott2019fairseq}. However, the motivation for this approach is primarily empirical. The second phenomenon we investigate is due to \citet{golatkar2019time} who show that in computer vision, applying WD only during say the first quarter of training is essentially as good as always applying it, and applying it after the first quarter is roughly as good as never applying it. We refer to these two schedules as early/late WD. We focus on the first quarter in this paper for concreteness but note that the same trend holds beyond exactly the first quarter. We will relate the observations of \citet{golatkar2019time} to the sharp/flat minima hypothesis of \citet{keskar2016large} which essentially states that the noise in SGD biases the network to flat minimizers which generalize well.

\textbf{Our Contributions.} Regarding observations of \cite{golatkar2019time}, we show that the network norm typically grows during the start, using WD early in training then ensures that the gradient steps are large relative to the weights throughout training. This has a regularizing effect, but traditional metrics of generalization \cite{keskar2016large} do not consistently capture this. We provide a scale-invariant metric to remedy this issue. We further demonstrate that dataset generalization properties significantly influence weight growth. Regarding observations due to \citet{loshchilov2017decoupled}, it is natural to believe that $l_2$ regularization and adaptivity are incompatible. We demonstrate that across RL (reinforcement learning) and NLP task, this is not the issue that decoupled weight decay solves, but instead that the gradients of $l_2$ terms can "drown" the gradient of the true objective function in the buffers of Adam \cite{kingma2014adam} (which stores estimates of the first and second-order moments of gradients). By decoupling the WD, the buffers are not shared between $l_2$ regularization and the true objective function, avoiding this mixing and facilitating hyperparameter tuning. We find no increase in absolute performance over sufficiently tuned WD, suggesting that hyperparameter stability rather than improved accuracy might be primarily responsible for decoupled WDs popularity. We conclude with lessons for practitioners regarding tuning and using WD.

\section{On the Temporal Dynamics of Weight Decay}

For investigating observations in \citet{golatkar2019time} we replicate their experimental setup with identical hyperparameters (listed in Table \ref{tab:hyperparams_dnn} in Appendix \ref{appendix:hyperparams}), training Resnet18 on Cifar10 and Resnet50 on Cifar100. We additionally provide experiments on tiny-imagenet \cite{tiny} using densenet 121 \cite{huang2017densely}. We consider this setting throughout the paper. In Figure \ref{fig:cifar_norms_over_time}, we show the weight norm and accuracy of networks with, without and with WD only after/before epoch 50 as per \cite{golatkar2019time}. We also consider a network with shuffled labels. We see that the norms of the network grow primarily at the start. By applying WD before epoch 50, we avoid the initial period of growth, and the norm stays low throughout training. Applying WD after epoch 50 results in many epochs before the norm reaches levels comparable to using WD throughout training.

\begin{figure*}[t!]
\centering
\includegraphics[width=0.95\textwidth]{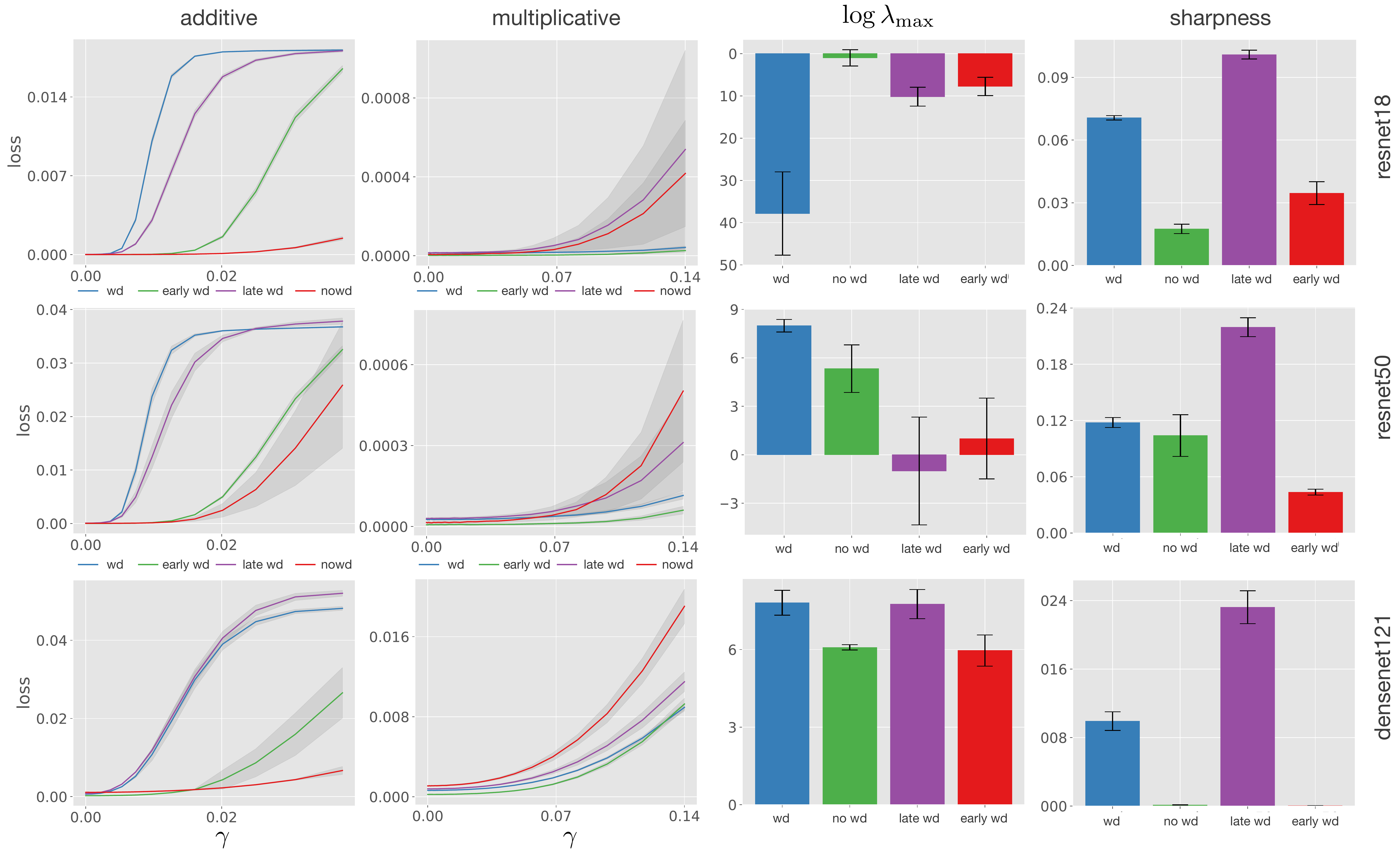}
\caption{The sharpness of networks, typically used as a proxy for generalization, using different WD schemes. We compare four metrics of sharpness: the largest hessian eigenvalues (compute via \citet{yao2019pyhessian}, measured logarithmically), the sharpness metric in \citet{keskar2016large} and additive/multiplicative perturbations. All metrics except multiplicative perturbations fail to consistently explain the differences in generalization of Figure \ref{fig:cifar_norms_over_time}. The loss under multiplicative perturbations increases when WD isn't used, suggesting that a sharp minima hypothesis might explain observations of \cite{golatkar2019time}. }
\label{fig:metrics_for_sharpness}
\end{figure*}

\noindent
\textbf{Early/Late Weight Decay and Generalization.} As per Figure \ref{fig:cifar_norms_over_time}, applying WD only at the start will ensure that the weight norm stays low during training. However, it is not clear why this would improve generalization; almost all network layers use batch normalization and are thus invariant under weight-rescaling. For a fixed learning rate and gradient, the effective change $\Delta w / w$ in the weights is smaller if the weights have a larger scale -- so decaying the weights increases the "effective" learning rate. A large learning rate and small batches typically have a regularizing effect as they induce noise into the training, and \citet{keskar2016large} have shown that large batches lead to sharp minimizers with poor generalization. Their explanation, which has garnered much attention \cite{li2018visualizing}, is essentially that networks with sharp minima generalize worse as they are more sensitive to the inherent shift between test/train surface. \citet{keskar2016large} uses the following metric of sharpness (with $\epsilon = 5\mathrm{e}{-4}$) for loss $L(\cdot)$ at a point $x$
\begin{equation} \label{eq:keskar_sharpness}
  \begin{gathered}
    \max_{y \in C_{\epsilon}} \frac{L(x + y) - L(x)}{L(x) + 1} \\
    C_{\epsilon} = \{ y \in R^n |  - \epsilon (1+|x_i|) \le y_i \le \epsilon (1+|x_i|) ] \}
  \end{gathered}
\end{equation}
In practice, $L(x+y)$ is maximized by first-order methods. Another common metric of sharpness is the largest eigenvalues of the Hessian \cite{iyer2020wide, dinh2017sharp}. In Figure \ref{fig:metrics_for_sharpness}, we plot these metrics and see that they typically give wrong or inconsistent results -- for example, the sharpness metric of \citet{keskar2016large} suggests that disabling WD yields flat minima which should generalize well -- the opposite of what we observe. We note that these metrics depend upon the scale of the network, and as per Figure \ref{fig:cifar_norms_over_time}, we know that networks without WD have larger norms. This motivates us to consider metrics of sharpness which are invariant under weigh scaling. We consider a simple scale-invariant metric -- multiplicative perturbations of the network weights 
\begin{figure*}[t!]
\centering
\includegraphics[width=0.9\textwidth]{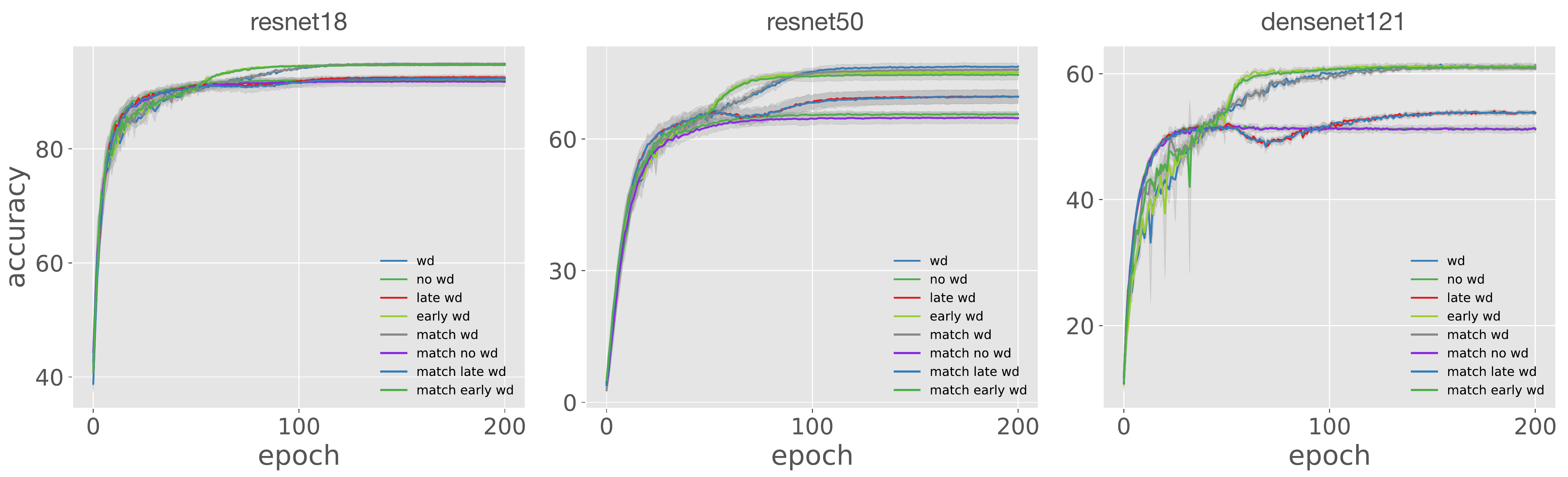}
\caption{Learning curves for DNNs trained without WD with weights scaled to match norms in Fig. \ref{fig:cifar_norms_over_time}, and the original learning curves that these DNNs are made to match. Scaling the weights roughly matches the performance of various WD schedules, suggesting that WD mediates the observations of \citet{golatkar2019time} through a simple scaling mechanism.}
\label{fig:matching_norm}
\end{figure*}
\begin{equation}
S(\gamma) = \mathbb{E} \big[ L ( w \odot (1 + \gamma \delta)) \big] \qquad \delta \sim N(0,I)
\end{equation}
\begin{figure*}[b!]
\centering
\includegraphics[width=0.9\textwidth]{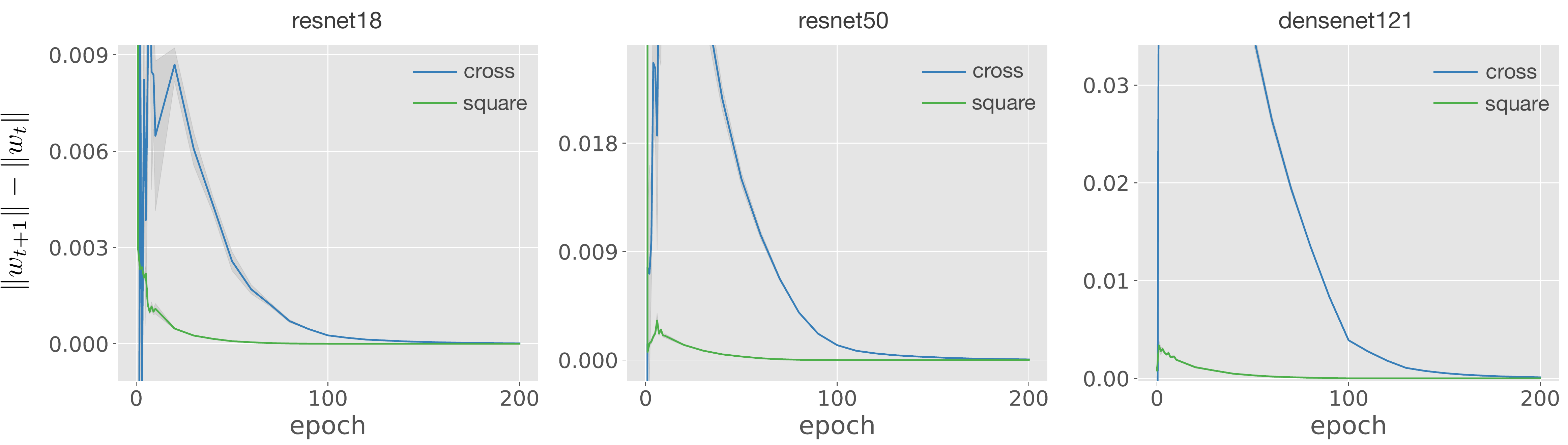}
\caption{The contributions to the change in weight norm for the square and cross terms, defined in \eqref{eq:define_cross_sq}. The cross term dominates and thus the norm grows primarily in the radial direction, scaling up subsets of the weights that align with the gradient.}
\label{fig:cross_vs_sq}
\end{figure*}
\begin{figure*}[t!]
\centering
\includegraphics[width=0.9\textwidth]{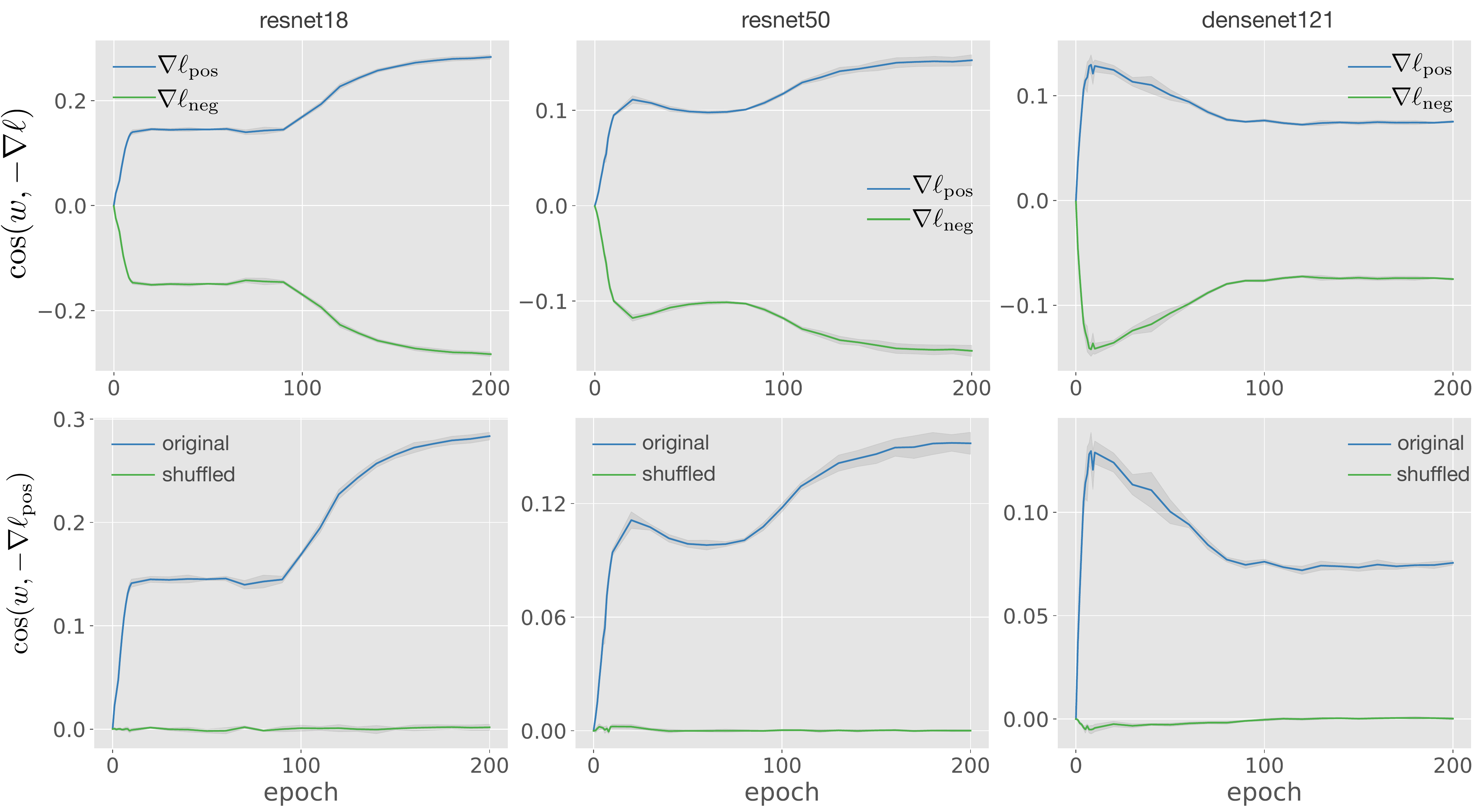}
\caption{\textit{(Top.)} We divide the cross entropy loss into two parts as per \eqref{eq:define_pos_neg}. The cosine between the weight vector $w$ and $-\nabla \ell_{\text{pos}}$ is positive whereas the cosine between $w$ and $-\nabla \ell_{\text{neg}}$ is negative. This suggests that network norm increases as subset of weights responsible for correct predictions grow in magnitude. \textit{(Bottom.)} $\cos(w, -\ell_{\text{pos}})$ with $\ell_{\text{pos}}$ defined as per \eqref{eq:define_pos_neg}. We see that for a network with shuffled labels the gradient barely points in the radial direction, which would lead to less growth as per Figure \ref{fig:cross_vs_sq}.}
\label{fig:pos_neg_std_shf}
\end{figure*}
That is, we scale each weight $w_i$ by $(1 + \gamma \delta_i)$ where $\delta_i$ is a standard normal variable, the intuition being that this metric measures how the loss changes to small multiplicative perturbations, say $\gamma \approx 0.1$. Note that this yields a metric similar to \cref{eq:keskar_sharpness}, with the chief difference being in how "small perturbations" are defined. The expectation is computed by sample averages. 

Figure \ref{fig:metrics_for_sharpness} illustrates this metric (referred to as "multiplicative"), showing that it gives results consistent with \citet{golatkar2019time}. We also show the results of an additive perturbation, which is analogous to multiplicative ones except that we take $w + \gamma \delta$. Note that it fails to capture generalization. Thus, we show that the explanation that sharp networks generalize worse can be applied to the empirical observations on early/late WD due to \citet{golatkar2019time} if one is careful regarding what is meant by sharpness. Our experiments also suggest that the effects of early/late WD are primarily mediated by modifying the effective learning rate $\Delta w / w$. To further solidify this hypothesis, we train networks without WD, but inspired by \citet{zhang2018three}, we manually scale the weights after each epoch to match the norm of another network trained with WD. Figure \ref{fig:matching_norm} shows that just scaling the weight norms is enough to achieve the results of \citet{golatkar2019time}. See \Cref{appendix:more_images} for further experiments without batch normalization and discussion.

\noindent
\textbf{On Causes for Weight Growth.} We have seen how applying WD early results in small weights throughout training on computer vision datasets, increasing the effective learning rate $\Delta w / w$. It is natural to believe that the network norm always grows during early parts of training, we here demonstrate that that's \textit{not} the case. Instead, the tendency of weight norms to grow is related to the dataset and its generalization properties. In Figure \ref{fig:cifar_norms_over_time}, we see that the weight norm of a network with shuffled labels stays almost constant during training. It's natural to wonder if the gradient norm might simply be smaller for shuffled labels, but this turns out to not be true, see Figure \ref{fig:gradient_norm} in Appendix \ref{appendix:more_images}. Indeed, the weights of networks trained on shuffled labels move significantly, just not in the radial direction, which would increase the weight norm, see Figure \ref{fig:dist_from_start} in Appendix \ref{appendix:more_images}. With shuffled labels, the dataset has the same images, but training on such a dataset will not generalize to a test set. This suggests that dataset generalization properties have an important influence on weight norms, which in turn modulates effective learning rates. To understand why the weights grow differently using original or shuffled labels, let us consider the weight norm change for an SGD update
\begin{equation}
\label{eq:define_cross_sq}
\| w_{t+1} \|^2 - \| w_t \|^2  = \underbrace{ \alpha^2 \| \nabla \ell_t \|^2 }_{\text{square term}} + \underbrace{ 2 \alpha \langle - \nabla \ell_t , w_t \rangle }_{\text{cross term}}
\end{equation}
There are two terms responsible for the increasing weights, a term that only depends on the gradient update and a cross term relating the direction of the gradient and the weights. In Figure \ref{fig:cross_vs_sq}, we illustrate how these two terms vary during optimization and find that the cross term is responsible for the lions share of the weight growth. Let us further divide the loss function into two parts representing the correct class and the normalization constant used in softmax 
\begin{equation}
\label{eq:define_pos_neg}
\ell_t (x) =  \frac{1}{|B|} \textstyle \sum_{i \in B} \underbrace{x_{i,\text{label}[i]}}_{\ell_{\text{pos}}} - \underbrace{\log \big( \textstyle \sum_j \exp(x_{ij}) \big) }_{\ell_{\text{neg}}}
\end{equation}
By linearity we of course have $\nabla \ell_t = \nabla \ell_{\text{pos}} + \nabla \ell_{\text{neg}}$ and thus $\langle \nabla \ell_t , w_t \rangle = \langle \nabla \ell_{\text{pos}} , w_t \rangle + \langle \nabla \ell_{\text{neg}} , w_t \rangle $. In Figure \ref{fig:pos_neg_std_shf} (top) we show how these two terms vary during optimization, observe that $-\nabla \ell_{\text{pos}}$ points along the weights while $-\nabla \ell_{\text{neg}}$ points away from the them. In light of Figure \ref{fig:cross_vs_sq}, we conclude that weight norms increases due to the gradient pointing roughly in the radial direction $w$, scaling up many weights $w_i$. Can this interpretation explain why shuffled labels lead to no weight growth? We first note that $\nabla \ell_{\text{neg}}$ is invariant under label permutations, and thus seek to look at $\nabla \ell_{\text{pos}}$. Figure \ref{fig:pos_neg_std_shf} (bottom) instead plots $\cos( -\nabla \ell_{\text{pos}} , w_t )$ for the standard network and a network with shuffled labels. There we see a striking difference, for the network trained on the original labels, the gradient typically points along the $w$ whereas gradients for networks using shuffled labels do not. To explain this, consider e.g. the last mini-batch $b$ we encounter in the first pass over the dataset. If we use the original labels, all images of e.g. dogs we have seen previously will likely push the network weights to increase the prediction probability of any dog pictures in batch $b$. Scaling up these weights will then decrease the loss. If we use shuffled labels however, simply scaling up network weights should not decrease the loss on batch $b$, since there is no generalization from dog pictures in previous batches. While an example with shuffled labels might seem artificial, the phenomenon of datasets influencing the weight norm growth happens in more natural settings such as RL. Figure \ref{fig:weights_over_time_atari} in Appendix \ref{appendix:more_images} shows that the network norm differ substantially between games when using identical hyperparameters for DQN \cite{mnih2015human}. Thus, if norm growth is dataset dependent, the observations of \cite{golatkar2019time} might only hold for datasets with good generalization.

\begin{figure*}[t!]
\centering
\includegraphics[width=0.95\textwidth]{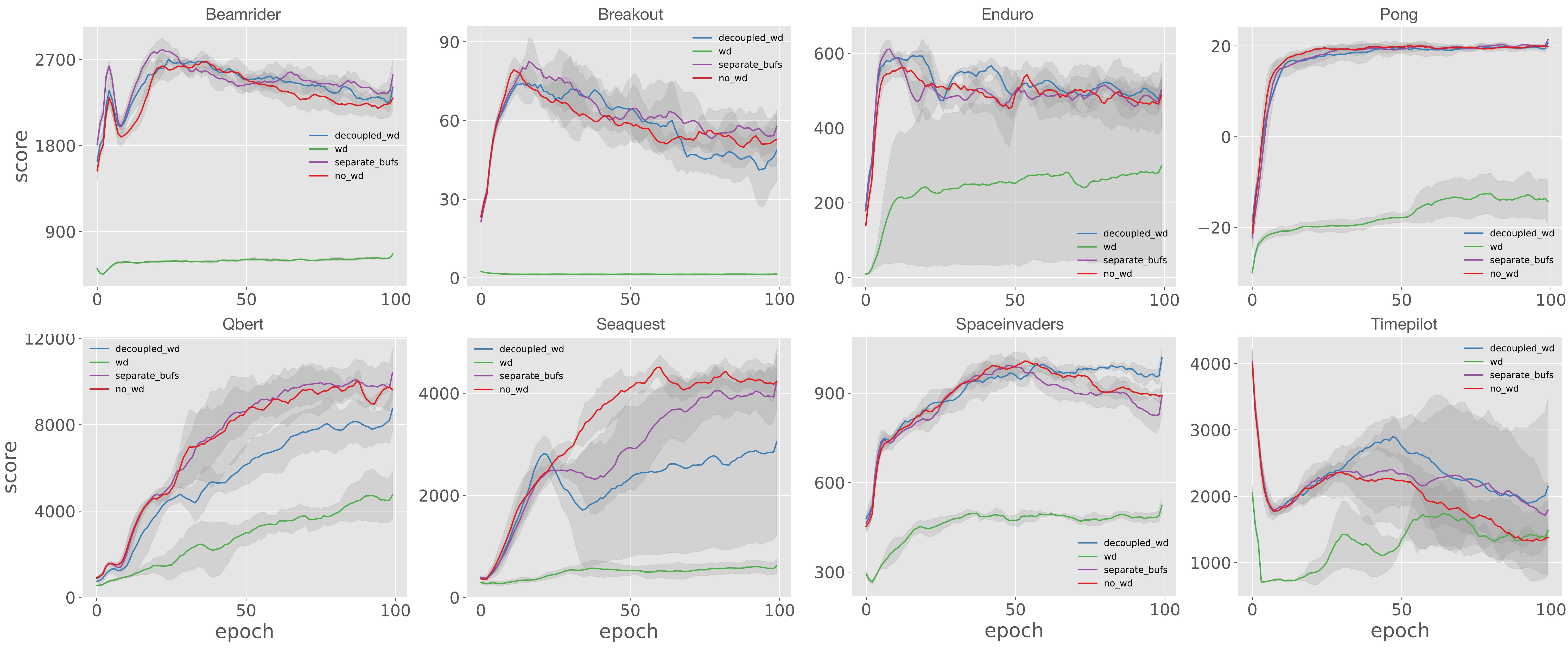}
\caption{Learning curves for various Atari games with WD ($\lambda = 0.0001$). We compare decoupled WD \cite{loshchilov2017decoupled}, original WD, no WD and an Adam variant with separate buffers for the WD and gradient signal. Original WD underperforms, whereas separating the buffers performs on par with decoupled WD. This suggests that the mixing of WD signal with the gradients, rather than the adaptivity itself, is responsible for the poor performance of normal WD in this setting.}
\label{fig:atari_scores_main}
\end{figure*}

\section{On Weight Decay for Adaptive Optimizers}
\label{sec:adaptive_optim}

\begin{figure*}[b!]
\centering
\includegraphics[width=\textwidth]{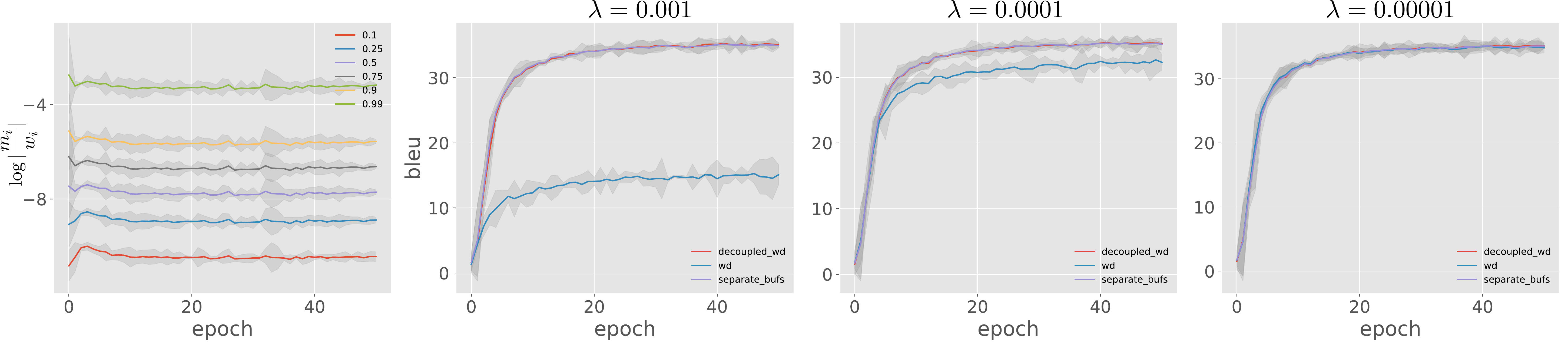}
\caption{The leftmost figure illustrates the quantiles of $\log |\frac{m_i}{w_i}|$ during training of a transformer \cite{vaswani2017attention}. They vary roughly two orders of magnitude, suggesting that the gradients for different parameters differ substantially. The three following figures illustrate the translation quality, measured in bleu, for three different values of $\lambda$ and three different weigh decay schemes. Standard WD underperforms unless $\lambda$ is taken small whereas separating the buffers matches decoupled WD.}
\label{fig:bleu_plus_q_new}
\end{figure*}

\citet{loshchilov2017decoupled} have proposed decoupled weight decay for adaptive optimizers, where one decays the weights by $(1-\alpha \lambda)$ instead of adding a $l_2$ penalty to the loss. We investigate this scheme in two contexts where adaptive optimizers are ubiquitous, NLP and RL. We first consider translation of the IWSLT'14 German to English dataset \cite{cettolo2014report} using transformer architectures \cite{vaswani2017attention} with code and default hyperparameters from the publicly available fairseq codebase \cite{ott2019fairseq}. We consider $\lambda \in \{ 1\mathrm{e}{-3}, 1\mathrm{e}{-4}, 1\mathrm{e}{-5}\}$, where the middle parameter is the default parameter used in fairseq, see Table \ref{tab:hyperparams_transformer} in Appendix \ref{appendix:hyperparams} for all hyperparameters. Secondly, we also consider the RL agent DQN \cite{mnih2015human}, using the publically available dopamine codebase \cite{castro18dopamine} with their default hyperparameters (see Appendix \ref{appendix:hyperparams}), trained on a handful of Atari games, most having been highlighted in previous work \cite{mnih2016asynchronous}. The three rightmost plots in Figure \ref{fig:bleu_plus_q_new} shows that for translation, WD under-performs decoupled WD except for the smallest value of $\lambda$. Similarly, Figure \ref{fig:atari_scores_main} shows that decoupled WD typically gives a sizable improvement in DQN, whereas WD can have a markedly deleterious effect on performance. To investigate why standard WD fails whereas decoupled WD succeeds, let us consider the buffer that Adam maintains to estimate the first moment of the gradient, which for loss function $\ell$ and $l_2$ loss is updated as
\begin{equation}
\label{eq:adam_mom_buf}
m^{t+1} \leftarrow (1- \beta_1) m^t + \beta_1 \nabla \ell + \beta_1 \lambda w_i
\end{equation}
In the leftmost plot of Figure \ref{fig:bleu_plus_q_new}, we consider the NLP task with WD turned off and show the distribution of the quantity $\frac{|m_i|}{|w_i|}$ for weight $w_i$, which roughly measures the strength of the gradient signal over the weight. The analogue illustration for DQN is found in Figure \ref{fig:atari_buf_over_weight} in the Appendix. In both these cases, we see that the distribution of absolute values of this quantity (plus $\epsilon$ for numerical stability) on a log scale, and see that 1) that the gradient signal is weak compared to the weight and 2) the scales are different by orders of magnitude for different weights. This means that the ratio between the gradients from the true objective function (the gradient signal) and the gradients from an $l_2$ penalty differs significantly between individual weights $w_i$. To avoid the WD signal dominating over the gradient signal in \eqref{eq:adam_mom_buf}, one would need to set $\lambda$ comparable to the smallest gradient signal. However, this might result in a very small value for the parameter with the largest gradient signal. Thus, effectively, the suitable ranges of $\lambda$ are dictated by the strength of the gradient signal. We can make this idea more precise with a scaling argument. For $l_2$ regularized Adam with weight $w_i$ and gradient strength $g_i$, equal to say the absolute value of an exponential average of the gradients, should shrink until we reach a steady state where $\lambda w_i \approx g_i$. If we assume that the ratio $m_i / (\sqrt{m_2} + \epsilon)$ of the Adam buffers are $\mathcal{O}(1)$ (i.e. the first moment and the square of the second moment are comparable), the effective update $\Delta w_i / w_i$ would be $\mathcal{O}( \alpha \lambda /g_i)$. For decoupled weight decay, the weights would shrink only until $\lambda w_i = \mathcal{O}(1)$ since Adam without WD is invariant under scaling of the gradient $g_i$. Thus the relative update $\Delta w_i / w_i$ would be $\mathcal{O}(\alpha \lambda)$. The important distinction is that the relative updates for decoupled WD only scales with hyperparameters we have control over, whereas for $l_2$ regularization it depends upon the dataset gradient signal which we cannot control, do not know a priori, and which might vary between parameters as per Figure \ref{fig:bleu_plus_q_new} and \ref{fig:atari_buf_over_weight} in \Cref{appendix:more_images}.

\begin{table*}[t!]
\caption{Average scores over three seeds for various Atari games and WD schemes. Standard WD consistently fails whereas most but not all games benefit from decoupled WD. Adding separate buffers, which stores estimates of the first and second order moments of the gradients, for the normal gradient and weight decay signal gives performance roughly matching decoupled WD. See Appendix \ref{appendix:more_images} for learning curves with standard deviations.}
\label{table:atari_scores}
\centering
\begin{tabular}{l|lllllllll}
\toprule
            & Beamr & Breako & Enduro& Pong & Qbert  & Seaq  & Spaceinv & Timep  & $\lambda $ \\ \midrule
orig        & 2293  & 52     & 488   &  19  & 9624   & 4230  & 892  & 1379 & 0.0 \\
decoupled   & 2843  & 77     &  545  &  22  & 11430  & 2956  & 1144 & 3136 & $1\mathrm{e}{-3}$ \\
WD          & 579   & 1      & 62    & -27  &  356   & 140   & 346  & 822  & $1\mathrm{e}{-3}$ \\
separated   & 3310  & 62     & 560   &  22  & 11369  & 1283  & 969  & 1784 & $1\mathrm{e}{-3}$ \\  \midrule
decoupled   & 2406  & 48     &  501  &  20  & 8733   & 3043  & 1018 & 2146 & $1\mathrm{e}{-4}$ \\
WD          & 666   & 1      & 298   & -14  &  4746  & 614   & 522  & 1480 & $1\mathrm{e}{-4}$ \\
separated   & 2535  & 57     & 502   &  21  & 10400  & 4231  & 889  & 1790 & $1\mathrm{e}{-4}$ \\ \midrule
decoupled   & 2481  & 58     &  517  &  21  & 10108  & 3358  & 956  & 699  & $1\mathrm{e}{-5}$ \\
WD          & 2375  & 114    & 230   &  16  &  11181 & 4856  & 1153 & 2821 & $1\mathrm{e}{-5}$ \\
separated   & 2367  & 47     & 565   &  21  & 8625   & 3055  & 846  & 2359 & $1\mathrm{e}{-5}$ \\ \midrule
WD          & 3257  & 89     & 589   & 21   &  11134 & 5095  & 1158 & 6707 & $1\mathrm{e}{-6}$ \\ 
WD          & 2947  & 49     & 475   & 21   &  9867  & 3953  & 748  & 4633 & $1\mathrm{e}{-7}$ \\ 
WD          & 2761  & 56     & 505   & 20   &  8384  & 4758  & 649  & 1927 & $1\mathrm{e}{-8}$ \\ 
\bottomrule
\end{tabular}
\end{table*}

This hypothesis predicts that the mixing of the WD signal and the gradient signal inside the Adam buffers is the important distinction between decoupled WD and $l_2$ regularization, and not the adaptivity itself. By allowing separate buffers in Adam (for both the first and second-order moments) for the gradients of the true objective and an $l_2$ penalty, we can investigate if the signal mixing indeed is the problem. We thus consider Adam with duplicate buffers $m_i, m_i', v_i, v_i'$ for the gradient and WD signal, see Appendix \ref{appendix:more_images} for a formal description. Note that as the gradient of the weight decay term appears in both the numerator and denominator of the buffers, the magnitude of the update for this scheme is invariant if the weight is rescaled, which is different from decoupled WD. Table \ref{table:atari_scores} and Figure \ref{fig:atari_scores_main} shows the result of this experiment for DQN, and we see that separating the buffers indeed leads to performance comparable to decoupled WD. Similarly, we can see for translation in the three rightmost plots of Figure \ref{fig:bleu_plus_q_new} that separating the buffers matches the performance of decoupled WD. We see that for sufficiently small $\lambda$, normal WD indeed does give an improvement in DQN. But what $\lambda$ is sufficiently small differs by at least an order of magnitude between games. Certain games (e.g., Enduro or Pong) requires $\lambda \le 1\mathrm{e}{-6}$ to give a comparable performance of no WD, whereas $1\mathrm{e}{-4}$ suffices for Timepilot. WD thus requires tuning $\lambda$, whereas decoupled WD is stable as observed by \cite{loshchilov2017decoupled}. We also note that WD sometimes outperforms decoupled WD, albeit with highly tuned $\lambda$, suggesting that the popularity of decoupled WD might be due to hyperparameter stability rather than absolute performance improvement. Indeed, state-of-the-art image classification network efficientnet \cite{tan2019efficientnet} does not use decoupled WD  for its adaptive optimizer.

\section{Discussion}

\noindent
\textbf{Related work.}  Weight decay has a long history as a regularizer in machine learning \cite{hinton1987learning, krogh1992simple}. The hypothesis that flat minima generalize is well-known \cite{hochreiter1997flat}, and has been proposed to explain why large batch learning fails to generalize \cite{keskar2016large}. The most prominent critique of the sharp-minima hypothesis comes from \citet{dinh2017sharp}, who proves that one can increase the sharpness of any given minima by reparametrizing the network. Similar criticism can be found in theoretical PAC-Bayes work \cite{tsuzuku2019normalized, rangamani2019scale, yi2019positively, neyshabur2017exploring}, that only provides experiments for large-vs-small batch sizes where standard sharpness metrics work well in practice. \citet{van2017l2} noted how WD would increase the relative size of gradient updates. This perspective was empirically substantiated in \citet{zhang2018three} who showed that this is the primary mechanism by WD improves generalization and also argues for the conditioning effect of decoupled WD. \citet{zhang2018three} is the only previous work on decoupled WD that we are aware of, whereas they primarily replicate experiments of \cite{loshchilov2017decoupled} and discuss the KFAC optimizer,  we focus on explaining \textit{why} decoupled WD improves hyperparameter stability. We do not know any work explaining the observations of \cite{golatkar2019time}.

\noindent
\textbf{Lessons for practitioners.} Our work points towards a few directly actionable insights. 1) Decoupled WD is useful in q-learning despite not being broadly used. However, different environments may need a separate WD parameter due to their different generalization behavior, suggesting the need for adaptive versions of WD. 2) Different datasets have norms that grow differently. Consequently, one should not näively transfer WD parameters between datasets, especially when they have different generalization properties. 3) If standard WD is used, one should pay close attention to the scale between gradient and WD signal when tuning $\lambda$. 4) Since the weight norm is the most important factor when using WD, one can apply WD only every few batches to save computational resources. A toy example of this on cifar10 is shown in Figure \ref{fig:stuttered} in Appendix \ref{appendix:more_images}, where WD is applied only every 128 batches with no performance cost. While WD rarely is the computational bottleneck, it cannot effectively be parallelized in a mirrored distributed strategy. Applied to more computationally intensive regularization such as \citet{xie2019adversarial}, this strategy might lead to substantial savings for larger models.

\textbf{Conclusions.} We have investigated recent empirical observations regarding WD. We observe that applying WD at the start increases the effective learning rate, which biases the network to less sharp minima. We also demonstrate that the primary distinction between decoupled weight decay and $l_2$ regularization is the sharing of buffers in Adam.

\section*{Acknowledgements}

This material is based upon work supported by the National Science Foundation under Grant Number CCF-1522054. This material is also based upon work supported by the Air Force Office of Scientific Research under award number FA9550-18-1-0136. This research is supported in part by the grants from Facebook, the National Science Foundation (III-1618134, III-1526012, IIS1149882, IIS-1724282, and TRIPODS- 1740822), the Office of Naval Research DOD (N00014- 17-1-2175), Bill and Melinda Gates Foundation. We are thankful for generous support by Zillow and SAP America Inc. We are also grateful from generous support from the TTS foundation. This work was partially supported by the Cornell Center for Materials Research with funding from the NSF MRSEC program (DMR-1719875).

\section*{Ethics and Broader Impact}

Our work extends the research community's understanding of weight decay, which is ubiquitously used in critical applications via neural networks. We do not perceive any entity to be directly put at a disadvantage or to be harmed due to any system failure. We do not believe that our research methods leverage biases in the data.

\bibliography{my_bib}

\begin{thebibliography}{32}
\providecommand{\natexlab}[1]{#1}
\providecommand{\url}[1]{\texttt{#1}}
\expandafter\ifx\csname urlstyle\endcsname\relax
  \providecommand{\doi}[1]{doi: #1}\else
  \providecommand{\doi}{doi: \begingroup \urlstyle{rm}\Url}\fi

\bibitem[Golatkar et~al.(2019)Golatkar, Achille, and Soatto]{golatkar2019time}
Aditya~Sharad Golatkar, Alessandro Achille, and Stefano Soatto.
\newblock Time matters in regularizing deep networks: Weight decay and data
  augmentation affect early learning dynamics, matter little near convergence.
\newblock In \emph{Advances in Neural Information Processing Systems}, pages
  10677--10687, 2019.

\bibitem[Tikhonov(1943)]{tikhonov1943stability}
Andrey~Nikolayevich Tikhonov.
\newblock On the stability of inverse problems.
\newblock In \emph{Dokl. Akad. Nauk SSSR}, volume~39, pages 195--198, 1943.

\bibitem[Hinton(1987)]{hinton1987learning}
Geoffrey~E Hinton.
\newblock Learning translation invariant recognition in a massively parallel
  networks.
\newblock In \emph{International Conference on Parallel Architectures and
  Languages Europe}, pages 1--13. Springer, 1987.

\bibitem[Tan and Le(2019)]{tan2019efficientnet}
Mingxing Tan and Quoc~V Le.
\newblock Efficientnet: Rethinking model scaling for convolutional neural
  networks.
\newblock \emph{arXiv preprint arXiv:1905.11946}, 2019.

\bibitem[Huang et~al.(2017)Huang, Liu, Van Der~Maaten, and
  Weinberger]{huang2017densely}
Gao Huang, Zhuang Liu, Laurens Van Der~Maaten, and Kilian~Q Weinberger.
\newblock Densely connected convolutional networks.
\newblock In \emph{Proceedings of the IEEE conference on computer vision and
  pattern recognition}, pages 4700--4708, 2017.

\bibitem[Ott et~al.(2019)Ott, Edunov, Baevski, Fan, Gross, Ng, Grangier, and
  Auli]{ott2019fairseq}
Myle Ott, Sergey Edunov, Alexei Baevski, Angela Fan, Sam Gross, Nathan Ng,
  David Grangier, and Michael Auli.
\newblock fairseq: A fast, extensible toolkit for sequence modeling.
\newblock In \emph{Proceedings of NAACL-HLT 2019: Demonstrations}, 2019.

\bibitem[Radford et~al.(2018)Radford, Narasimhan, Salimans, and
  Sutskever]{radford2018improving}
Alec Radford, Karthik Narasimhan, Tim Salimans, and Ilya Sutskever.
\newblock Improving language understanding by generative pre-training.
\newblock \emph{URL https://s3-us-west-2. amazonaws.
  com/openai-assets/researchcovers/languageunsupervised/language understanding
  paper. pdf}, 2018.

\bibitem[Loshchilov and Hutter(2017)]{loshchilov2017decoupled}
Ilya Loshchilov and Frank Hutter.
\newblock Decoupled weight decay regularization.
\newblock \emph{arXiv preprint arXiv:1711.05101}, 2017.

\bibitem[Kingma and Ba(2014)]{kingma2014adam}
Diederik~P Kingma and Jimmy Ba.
\newblock Adam: A method for stochastic optimization.
\newblock \emph{arXiv preprint arXiv:1412.6980}, 2014.

\bibitem[Wang et~al.(2018)Wang, Yuan, Li, Yu, and Jian]{wang2018sface}
Jianfeng Wang, Ye~Yuan, Boxun Li, Gang Yu, and Sun Jian.
\newblock Sface: An efficient network for face detection in large scale
  variations.
\newblock \emph{arXiv preprint arXiv:1804.06559}, 2018.

\bibitem[Carion et~al.(2020)Carion, Massa, Synnaeve, Usunier, Kirillov, and
  Zagoruyko]{carion2020end}
Nicolas Carion, Francisco Massa, Gabriel Synnaeve, Nicolas Usunier, Alexander
  Kirillov, and Sergey Zagoruyko.
\newblock End-to-end object detection with transformers.
\newblock \emph{arXiv preprint arXiv:2005.12872}, 2020.

\bibitem[Liu et~al.(2020)Liu, Blessing, Wood, and Lim]{liu2020crisisbert}
Junhua Liu, Trisha Singhal~Lucienne Blessing, Kristin~L Wood, and Kwan~Hui Lim.
\newblock Crisisbert: Robust transformer for crisis classification and
  contextual crisis embedding.
\newblock \emph{arXiv preprint arXiv:2005.06627}, 2020.

\bibitem[Keskar et~al.(2016)Keskar, Mudigere, Nocedal, Smelyanskiy, and
  Tang]{keskar2016large}
Nitish~Shirish Keskar, Dheevatsa Mudigere, Jorge Nocedal, Mikhail Smelyanskiy,
  and Ping Tak~Peter Tang.
\newblock On large-batch training for deep learning: Generalization gap and
  sharp minima.
\newblock \emph{arXiv preprint arXiv:1609.04836}, 2016.

\bibitem[Karpathy et~al.(2017)Karpathy, Li, and Johnson]{tiny}
Andrej Karpathy, Fei-Fei Li, and Justin Johnson.
\newblock {BWorld Robot Control Software}.
\newblock \url{https://tiny-imagenet.herokuapp.com/}, 2017.

\bibitem[Yao et~al.(2019)Yao, Gholami, Keutzer, and Mahoney]{yao2019pyhessian}
Zhewei Yao, Amir Gholami, Kurt Keutzer, and Michael Mahoney.
\newblock Pyhessian: Neural networks through the lens of the hessian.
\newblock \emph{arXiv preprint arXiv:1912.07145}, 2019.

\bibitem[Li et~al.(2018)Li, Xu, Taylor, Studer, and
  Goldstein]{li2018visualizing}
Hao Li, Zheng Xu, Gavin Taylor, Christoph Studer, and Tom Goldstein.
\newblock Visualizing the loss landscape of neural nets.
\newblock In \emph{Advances in Neural Information Processing Systems}, pages
  6389--6399, 2018.

\bibitem[Iyer et~al.(2020)Iyer, Thejas, Kwatra, Ramjee, and
  Sivathanu]{iyer2020wide}
Nikhil Iyer, V~Thejas, Nipun Kwatra, Ramachandran Ramjee, and Muthian
  Sivathanu.
\newblock Wide-minima density hypothesis and the explore-exploit learning rate
  schedule.
\newblock \emph{arXiv preprint arXiv:2003.03977}, 2020.

\bibitem[Dinh et~al.(2017)Dinh, Pascanu, Bengio, and Bengio]{dinh2017sharp}
Laurent Dinh, Razvan Pascanu, Samy Bengio, and Yoshua Bengio.
\newblock Sharp minima can generalize for deep nets.
\newblock In \emph{Proceedings of the 34th International Conference on Machine
  Learning-Volume 70}, pages 1019--1028. JMLR. org, 2017.

\bibitem[Zhang et~al.(2018)Zhang, Wang, Xu, and Grosse]{zhang2018three}
Guodong Zhang, Chaoqi Wang, Bowen Xu, and Roger Grosse.
\newblock Three mechanisms of weight decay regularization.
\newblock \emph{arXiv preprint arXiv:1810.12281}, 2018.

\bibitem[Mnih et~al.(2015)Mnih, Kavukcuoglu, Silver, Rusu, Veness, Bellemare,
  Graves, Riedmiller, Fidjeland, Ostrovski, et~al.]{mnih2015human}
Volodymyr Mnih, Koray Kavukcuoglu, David Silver, Andrei~A Rusu, Joel Veness,
  Marc~G Bellemare, Alex Graves, Martin Riedmiller, Andreas~K Fidjeland, Georg
  Ostrovski, et~al.
\newblock Human-level control through deep reinforcement learning.
\newblock \emph{Nature}, 518\penalty0 (7540):\penalty0 529--533, 2015.

\bibitem[Cettolo et~al.(2014)Cettolo, Niehues, St{\"u}ker, Bentivogli, and
  Federico]{cettolo2014report}
Mauro Cettolo, Jan Niehues, Sebastian St{\"u}ker, Luisa Bentivogli, and
  Marcello Federico.
\newblock Report on the 11th iwslt evaluation campaign, iwslt 2014.
\newblock In \emph{Proceedings of the International Workshop on Spoken Language
  Translation, Hanoi, Vietnam}, volume~57, 2014.

\bibitem[Vaswani et~al.(2017)Vaswani, Shazeer, Parmar, Uszkoreit, Jones, Gomez,
  Kaiser, and Polosukhin]{vaswani2017attention}
Ashish Vaswani, Noam Shazeer, Niki Parmar, Jakob Uszkoreit, Llion Jones,
  Aidan~N Gomez, {\L}ukasz Kaiser, and Illia Polosukhin.
\newblock Attention is all you need.
\newblock In \emph{Advances in neural information processing systems}, pages
  5998--6008, 2017.

\bibitem[Castro et~al.(2018)Castro, Moitra, Gelada, Kumar, and
  Bellemare]{castro18dopamine}
Pablo~Samuel Castro, Subhodeep Moitra, Carles Gelada, Saurabh Kumar, and
  Marc~G. Bellemare.
\newblock Dopamine: {A} {R}esearch {F}ramework for {D}eep {R}einforcement
  {L}earning.
\newblock 2018.
\newblock URL \url{http://arxiv.org/abs/1812.06110}.

\bibitem[Mnih et~al.(2016)Mnih, Badia, Mirza, Graves, Lillicrap, Harley,
  Silver, and Kavukcuoglu]{mnih2016asynchronous}
Volodymyr Mnih, Adria~Puigdomenech Badia, Mehdi Mirza, Alex Graves, Timothy
  Lillicrap, Tim Harley, David Silver, and Koray Kavukcuoglu.
\newblock Asynchronous methods for deep reinforcement learning.
\newblock In \emph{International conference on machine learning}, pages
  1928--1937, 2016.

\bibitem[Krogh and Hertz(1992)]{krogh1992simple}
Anders Krogh and John~A Hertz.
\newblock A simple weight decay can improve generalization.
\newblock In \emph{Advances in neural information processing systems}, pages
  950--957, 1992.

\bibitem[Hochreiter and Schmidhuber(1997)]{hochreiter1997flat}
Sepp Hochreiter and J{\"u}rgen Schmidhuber.
\newblock Flat minima.
\newblock \emph{Neural Computation}, 9\penalty0 (1):\penalty0 1--42, 1997.

\bibitem[Tsuzuku et~al.(2019)Tsuzuku, Sato, and
  Sugiyama]{tsuzuku2019normalized}
Yusuke Tsuzuku, Issei Sato, and Masashi Sugiyama.
\newblock Normalized flat minima: Exploring scale invariant definition of flat
  minima for neural networks using pac-bayesian analysis.
\newblock \emph{arXiv preprint arXiv:1901.04653}, 2019.

\bibitem[Rangamani et~al.(2019)Rangamani, Nguyen, Kumar, Phan, Chin, and
  Tran]{rangamani2019scale}
Akshay Rangamani, Nam~H Nguyen, Abhishek Kumar, Dzung Phan, Sang~H Chin, and
  Trac~D Tran.
\newblock A scale invariant flatness measure for deep network minima.
\newblock \emph{arXiv preprint arXiv:1902.02434}, 2019.

\bibitem[Yi et~al.(2019)Yi, Meng, Chen, Ma, and Liu]{yi2019positively}
Mingyang Yi, Qi~Meng, Wei Chen, Zhi-ming Ma, and Tie-Yan Liu.
\newblock Positively scale-invariant flatness of relu neural networks.
\newblock \emph{arXiv preprint arXiv:1903.02237}, 2019.

\bibitem[Neyshabur et~al.(2017)Neyshabur, Bhojanapalli, McAllester, and
  Srebro]{neyshabur2017exploring}
Behnam Neyshabur, Srinadh Bhojanapalli, David McAllester, and Nati Srebro.
\newblock Exploring generalization in deep learning.
\newblock In \emph{Advances in Neural Information Processing Systems}, pages
  5947--5956, 2017.

\bibitem[Van~Laarhoven(2017)]{van2017l2}
Twan Van~Laarhoven.
\newblock L2 regularization versus batch and weight normalization.
\newblock \emph{arXiv preprint arXiv:1706.05350}, 2017.

\bibitem[Xie et~al.(2019)Xie, Tan, Gong, Wang, Yuille, and
  Le]{xie2019adversarial}
Cihang Xie, Mingxing Tan, Boqing Gong, Jiang Wang, Alan Yuille, and Quoc~V Le.
\newblock Adversarial examples improve image recognition.
\newblock \emph{arXiv preprint arXiv:1911.09665}, 2019.

\end{thebibliography}
\newpage
\appendix
\onecolumn
\setcounter{secnumdepth}{3}
\section{Hyperparameters}
\label{appendix:hyperparams}

\begin{table}[h]
\caption{Hyper-parameters for training DNNs. These are used for all experiments except shuffled labels on cifa10, where they don't give $100\%$ train score. For this reason we use initial lr of $0.01$ for this experiment.}
\label{tab:hyperparams_dnn}
\begin{center}
    \begin{tabular}{ l | l }
    \hline
    Parameter & Value \\ \hline
    init. learning rate & 0.1 \\ 
    learning rate decay per step & 0.97 \\ 
    SGD momentum & 0.9 \\ 
    batch size & 128 \\ 
    randomized crop & pad 4, crop 32 \\
    horizontal flipping & True \\
    loss & cross-entropy \\
    weight decay & 0.0005 \\
    loss & cross-entropy \\ \hline
    \end{tabular}
\end{center}
\end{table}

\begin{table}[h]
\caption{Hyper-parameters used for DQN.}
\label{tab:hyperparams_dqn}
\begin{center}
    \begin{tabular}{ l | l }
    \hline
    Parameter & Value \\ \hline
    $\gamma$ & 0.9 \\ 
    $\epsilon_{\text{train}}$ & 0.01 \\
    $\epsilon_{\text{test}}$ & 0.001 \\
    learning rate & 0.00025 \\
    $\beta_1, \beta_2$ & 0.9, 0.999 \\
    $\epsilon_{\text{adam}}$ & 0.00000001 \\
    sticky actions & True \\ 
    batch size & 32 \\ 
    replay buffer capacity & 1,000,000 \\ 
    steps per epoch & 250,000 \\
    max steps per episode & 2,700 \\ \hline
    \end{tabular}
\end{center}
\end{table}

\begin{table}[h]
\caption{Hyper-parameters used for Transformers \cite{vaswani2017attention}. The architecture is "transformer-iwslt-de-en" of Fairseq.}
\label{tab:hyperparams_transformer}
\begin{center}
    \begin{tabular}{ l | l }
    \hline
    Parameter & Value \\ \hline
    learning rate & 0.0005 \\
    $\beta_1, \beta_2$ & 0.9, 0.98 \\
    $\epsilon_{\text{adam}}$ & 0.00000001 \\
    batch size & 32 \\
    label smoothing & 0.1 \\ 
    dropout probability & 0.3 \\ 
    max-tokens & 4096 \\ \hline
    \end{tabular}
\end{center}
\end{table}

\section{Further Experiments}
\label{appendix:more_images}

We here perform various experiments referred to in the main text. Additionally, we perform experiments with networks not using batch normalization at the suggestion of a NeurIPS reviewer. Regarding the parameter choice of $\gamma$, we have observed that for large $\gamma$, the multiplicative metric sometimes does not distinguish between early and late weight decay for the densenet. Large $\gamma$ does not correspond to small perturbations around a minimizer, and $\gamma \approx 0.1$ gives good results across our experiments and represents changing the weights roughly $10 \%$. We thus take $\gamma = 0.1$ to be the default parameter of our metric, and note that e.g. \citet{keskar2016large} also uses parameters to define their sharpness metric.

\begin{figure*}[b!]
\centering
\includegraphics[width=\textwidth]{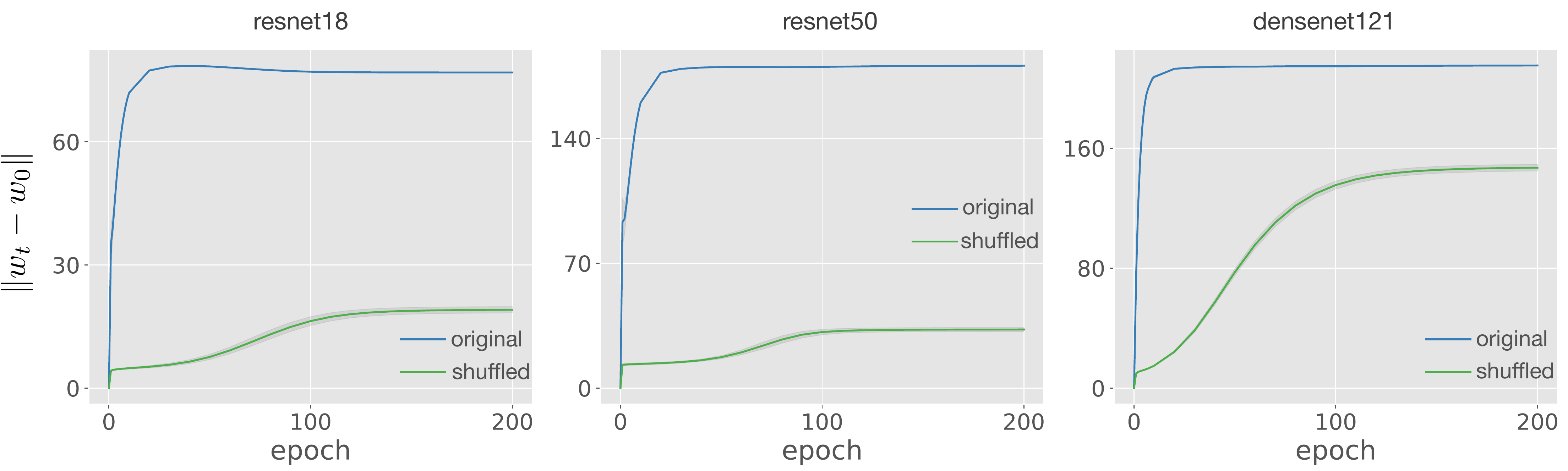}
\caption{The distance from the starting point for a standard network and one with shuffled labels and no wd. The distance is large compared to the change in norm of Figure \ref{fig:cifar_norms_over_time} for networks with shuffled labels, suggesting that the direction of movement is important and not only the absolute distance.}
\label{fig:dist_from_start}
\end{figure*}

\begin{figure*}
\centering
\includegraphics[width=\textwidth]{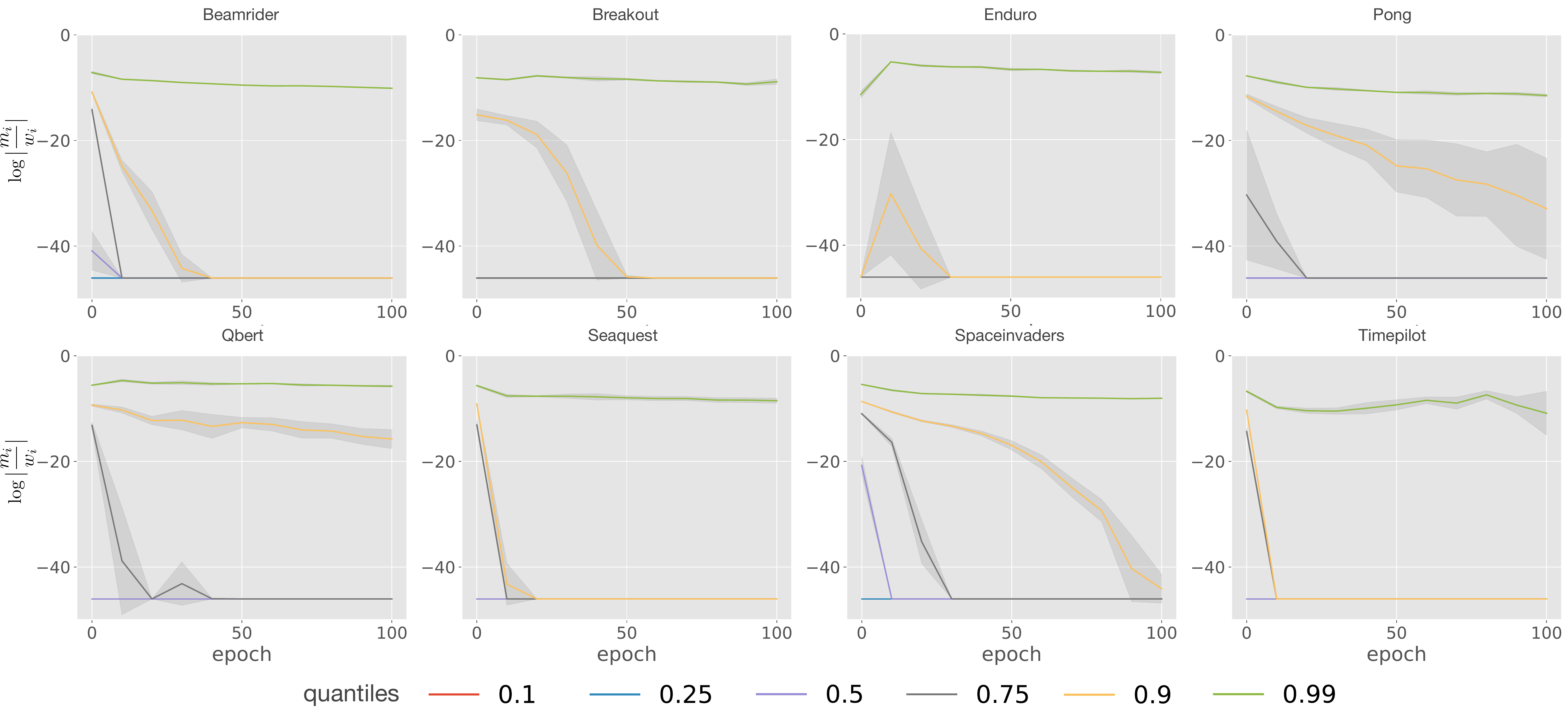}
\caption{Quantiles of $ \log \big| \frac{m_i}{w_i} \big|$ for weight $w_i$ with buffer $m_i$ as a function of time for DQN \cite{mnih2015human} trained on various Atari games. The gradient signal is small compared to the weights, but the ratio differs dramatically between parameters.}
\label{fig:atari_buf_over_weight}
\end{figure*}

\begin{figure}[h]
\centering
\includegraphics[width=\textwidth]{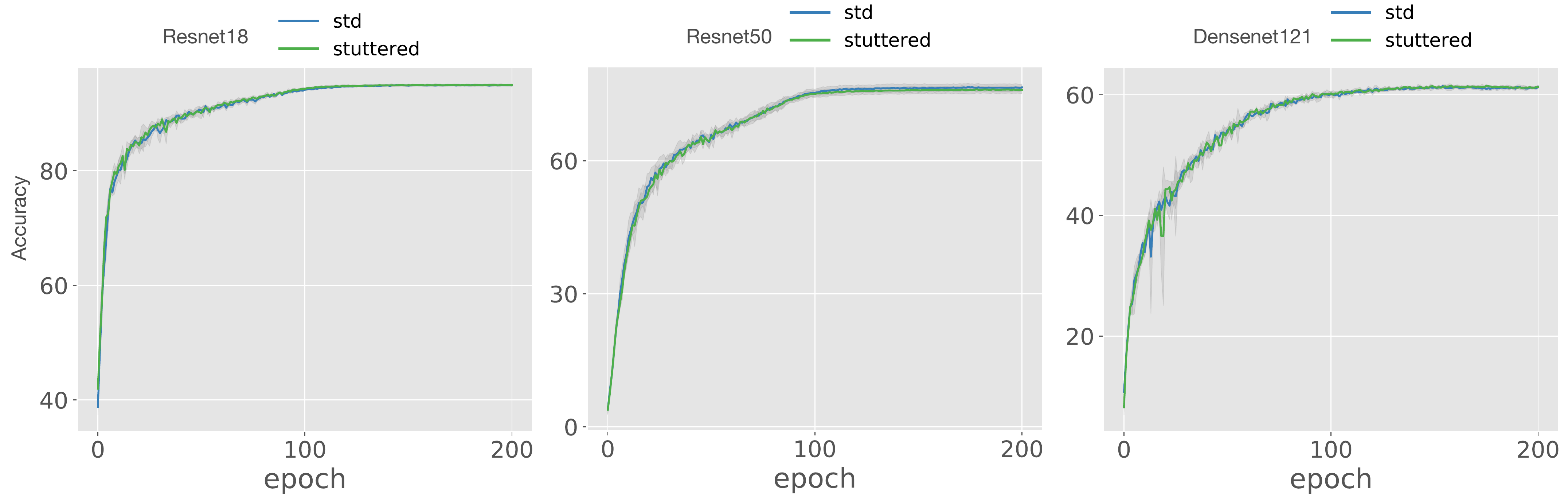}
\caption{The effect of applying wd only every $128$ SGD update, which we call stuttered. The final accuracy is unchanged across architectures and datasets, presenting opportunities for computational savings.}
\label{fig:stuttered}
\end{figure}

\begin{algorithm}[H]
\SetAlgoLined
 $\hat{v}_t \leftarrow 0 $\\
 $\hat{m}_t \leftarrow 0 $\\
 \While{$\theta_t$ not converged}{
   $\hat{m}_t \leftarrow \beta_1 \hat{m}_{t-1} + (1-\beta_1) g$ \\
   $\hat{v}_t \leftarrow \beta_2 \hat{v}_{t-1} + (1-\beta_2) g^2$ \\
    $\hat{m}_t' \leftarrow \beta_1 \hat{m}_{t-1} + (1-\beta_1) \theta_t$ \\
   $\hat{v}_t' \leftarrow \beta_2 \hat{v}_{t-1} + (1-\beta_2) \theta_t^2$ \\
   $\theta_{t} \leftarrow \theta_{t-1} - \alpha \hat{m}_t / (\sqrt{\hat{v}_t} + \epsilon)$ \\
    $\theta_{t} \leftarrow \theta_{t-1} - \alpha \lambda \hat{m}_t' / (\sqrt{\hat{v}_t'} + \epsilon)$
 }
\caption{Buffers for weight decay. A separate Adam optimizer with separate buffers is used for the gradients. Bias correction omitted for clarity, see \citet{kingma2014adam}.}
\end{algorithm}

\begin{figure}[h]
\centering
\includegraphics[width=\textwidth]{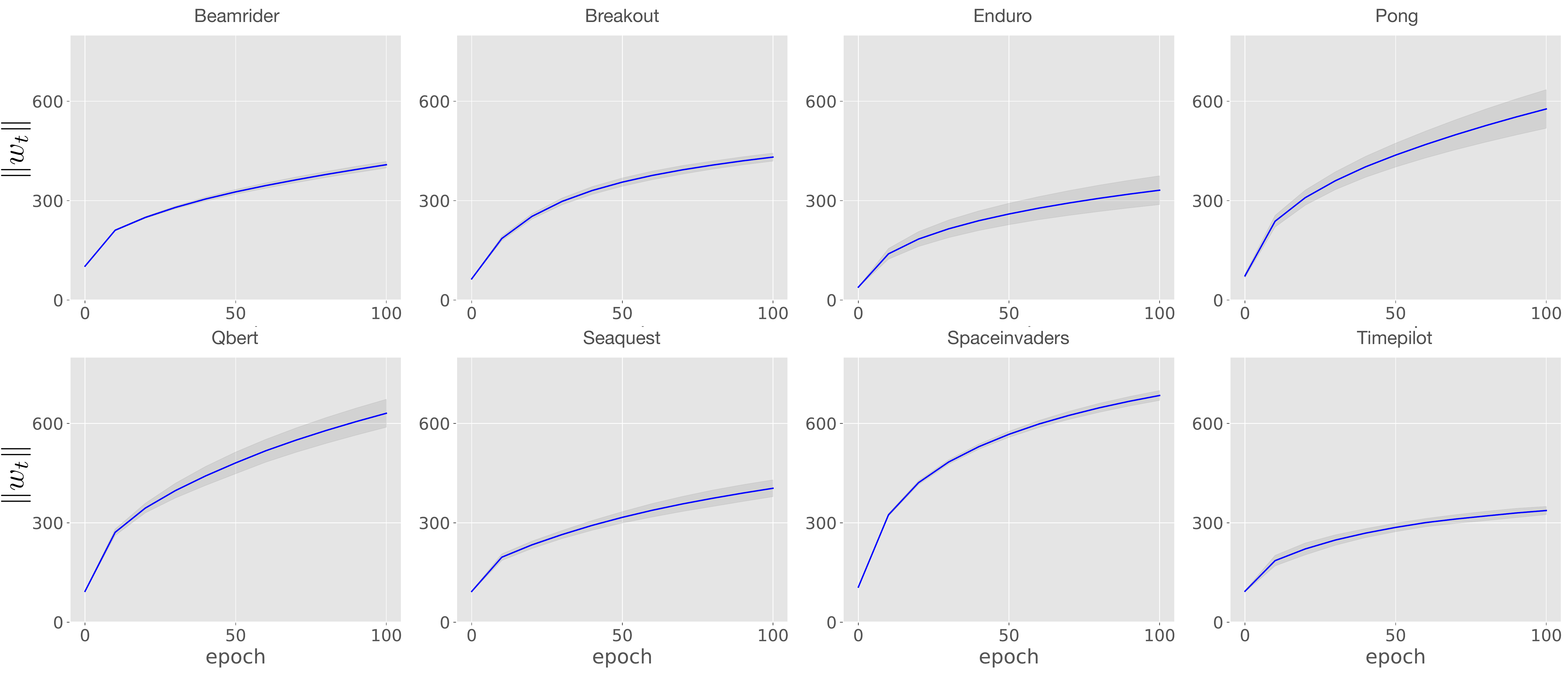}
\label{fig:diff_weights_atari}
\caption{Norms of weights of DQN for a couple of games from the Atari suite, all trained with identical hyperparameters. Norm growth differs not only for shuffled the labels, but also across more realistic datasets. One sees significant differences in the learning dynamics compared to image classification, the non-stationary nature of the dataset in RL might be partly responsible for this difference.}
\label{fig:weights_over_time_atari}
\end{figure}

\begin{figure}[h]
\centering
\includegraphics[width=\textwidth]{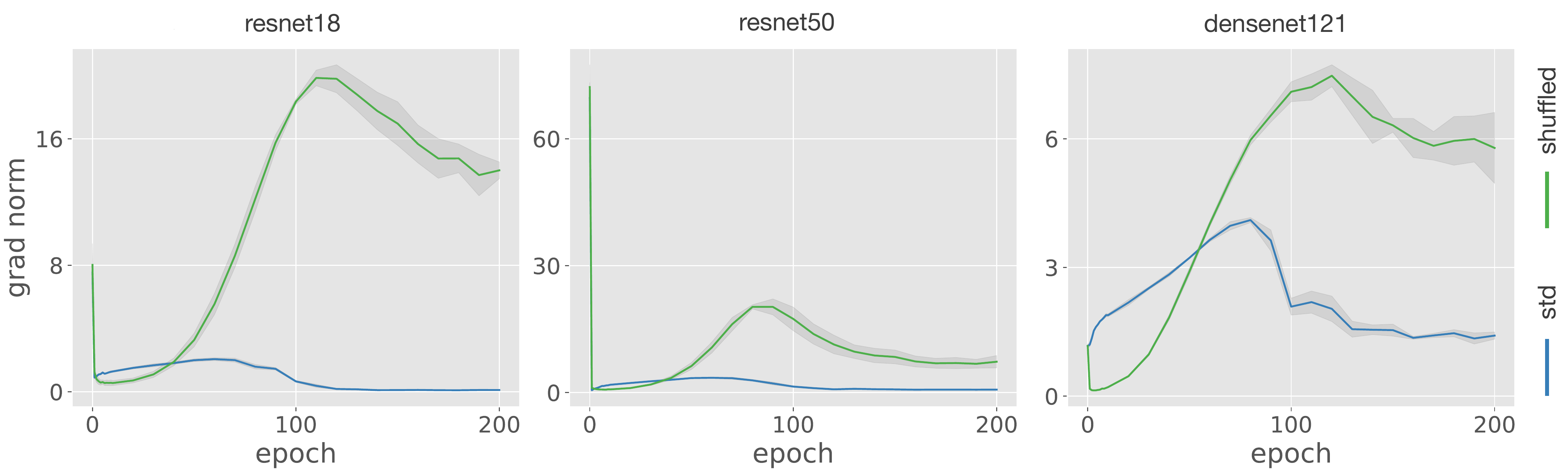}
\caption{The gradient norm during training for a standard network, and a network with shuffled labels and no weight decay. The gradients have roughly the same order of magnitude, and later in the training shuffled labels typically leads to larger gradients.}
\label{fig:gradient_norm}
\end{figure}

\begin{figure}[h]
\centering
\includegraphics[width=\textwidth]{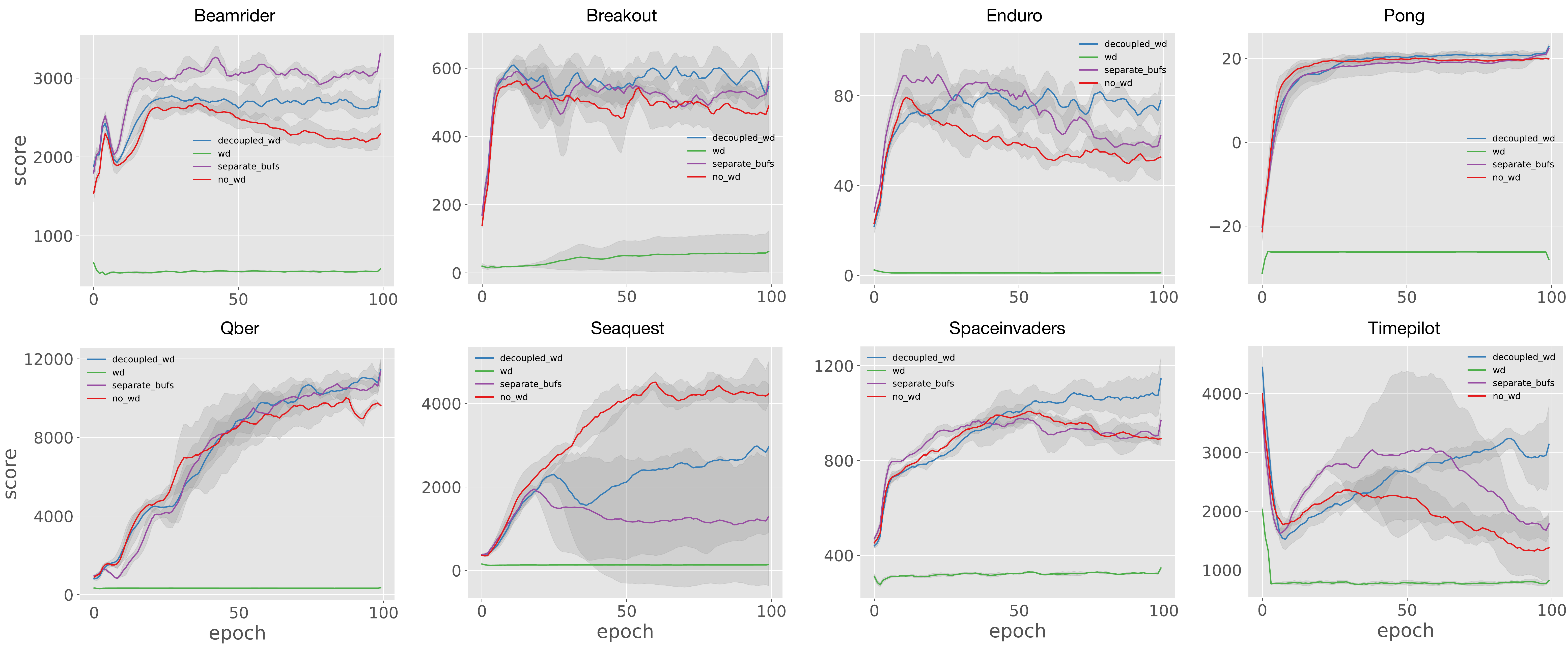}
\caption{Learning curves for various Atari games for $\lambda = 0.001$.}
\end{figure}

\begin{figure}[h]
\centering
\includegraphics[width=\textwidth]{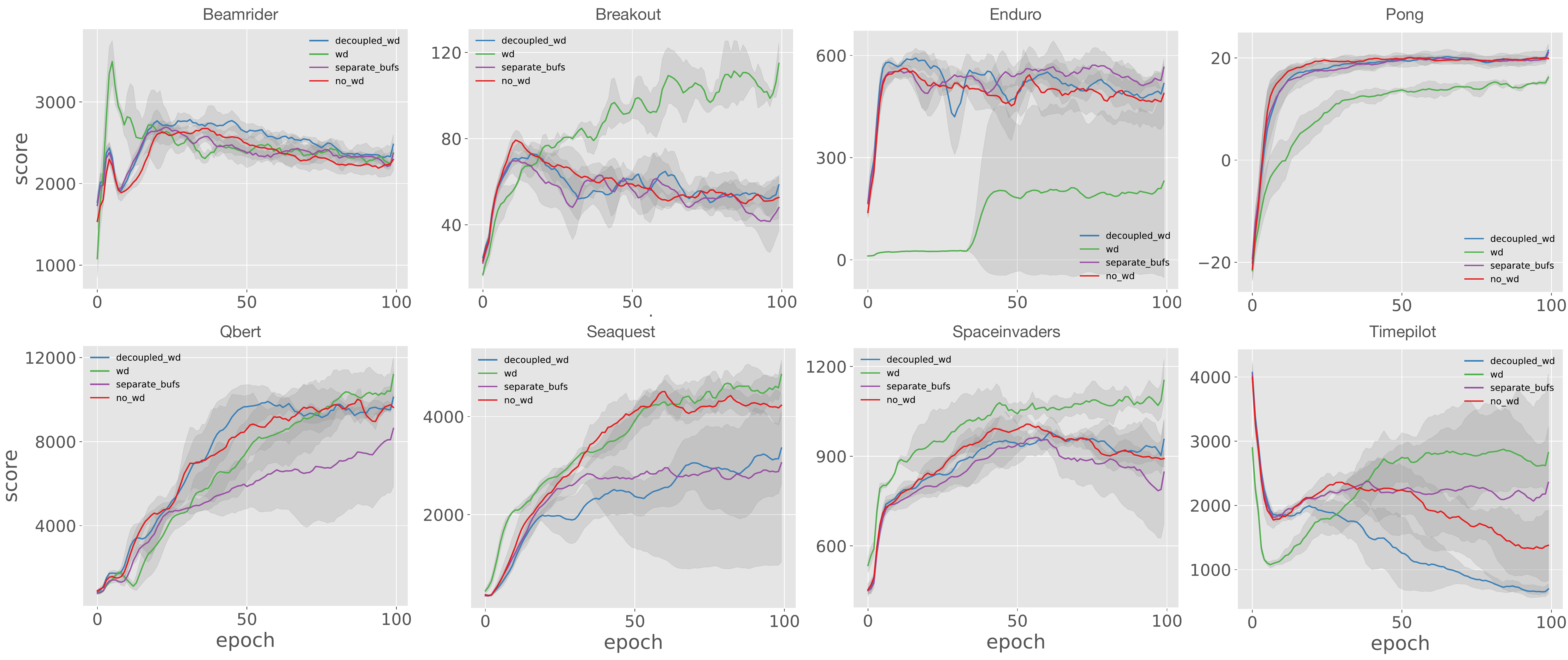}
\caption{Learning curves for various Atari games with $\lambda = 0.00001$.}
\end{figure}

\begin{figure}[h]
\centering
\includegraphics[width=\textwidth]{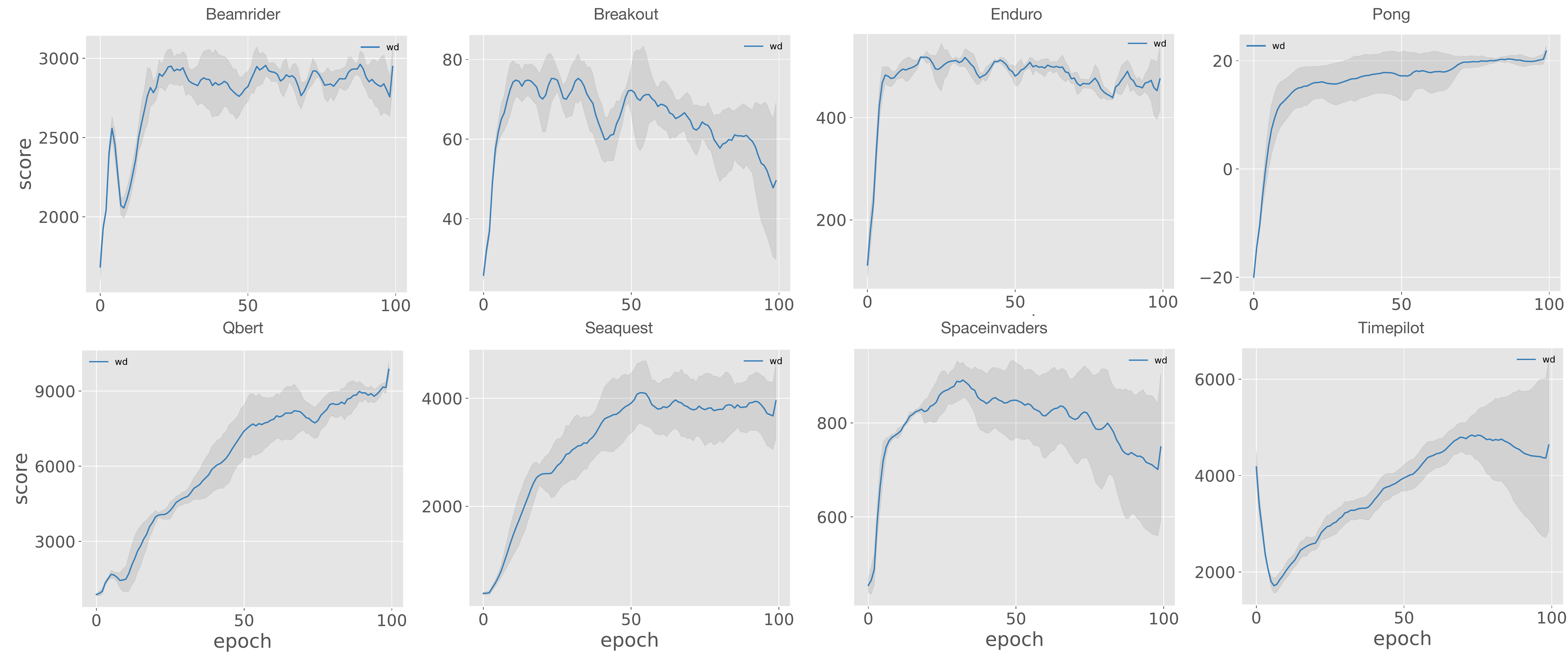}
\caption{Learning curves for various Atari games with $\lambda = 0.0000001.$}
\end{figure}

\begin{figure}[h]
\centering
\includegraphics[width=\textwidth]{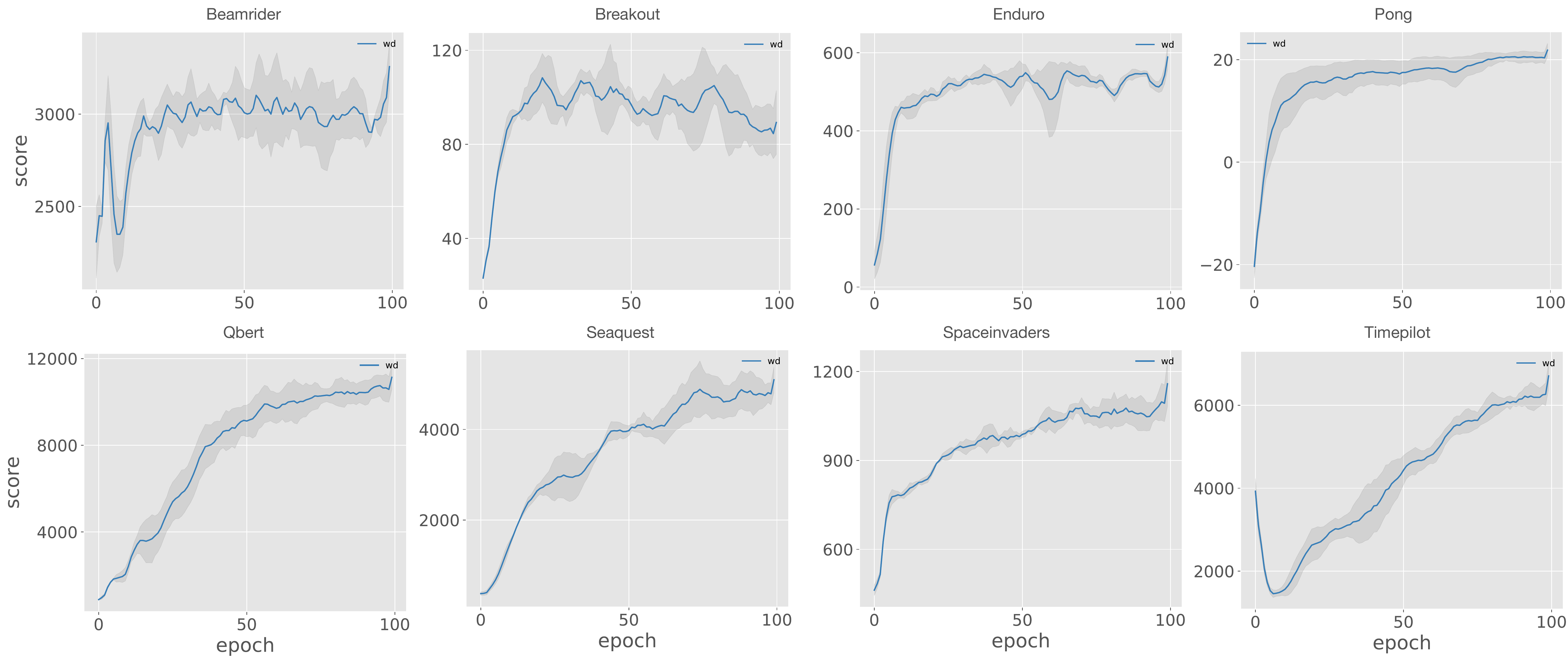}
\caption{Learning curves for various Atari games with $\lambda = 0.000001$.}
\end{figure}

\begin{figure}[h]
\centering
\includegraphics[width=\textwidth]{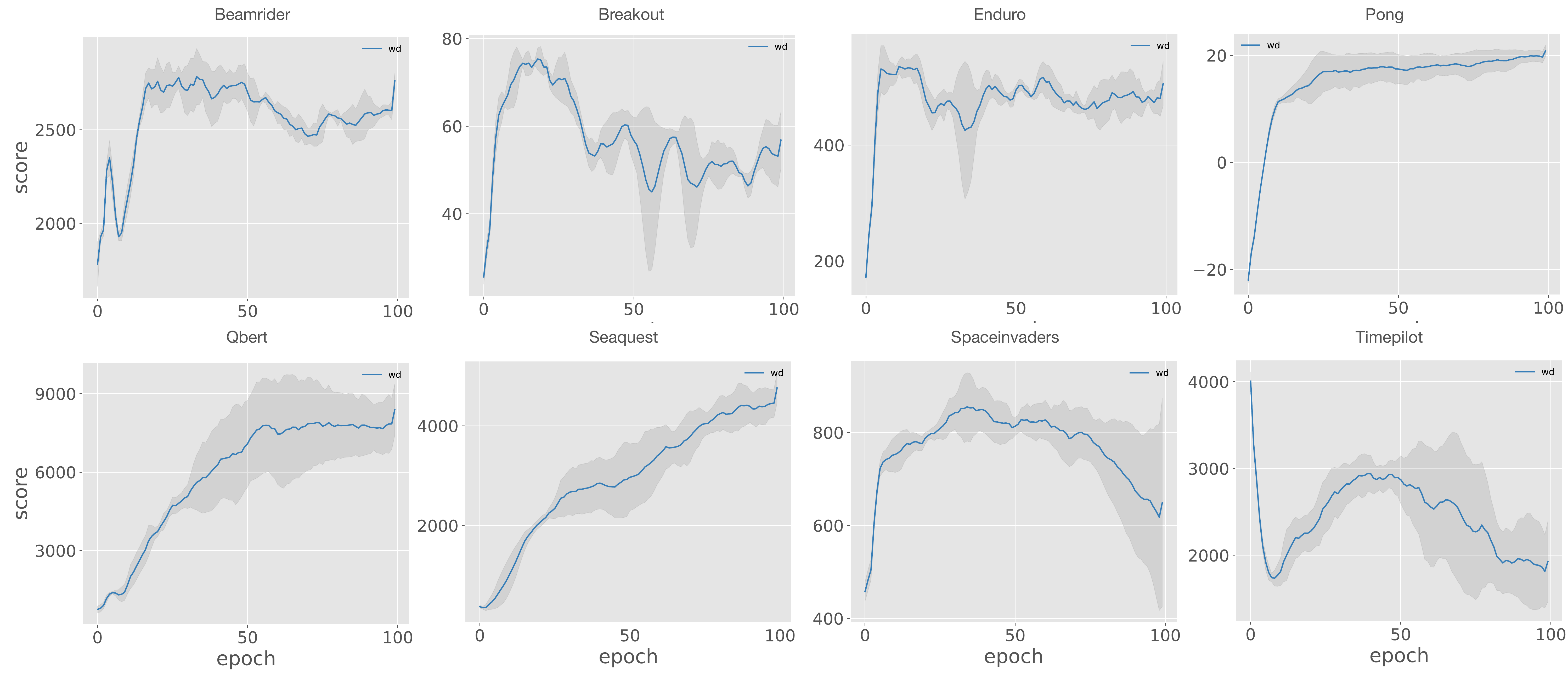}
\caption{Learning curves for various Atari games with $\lambda = 0.00000001$.}
\end{figure}

\begin{figure}[h]
\centering
\includegraphics[width=\textwidth]{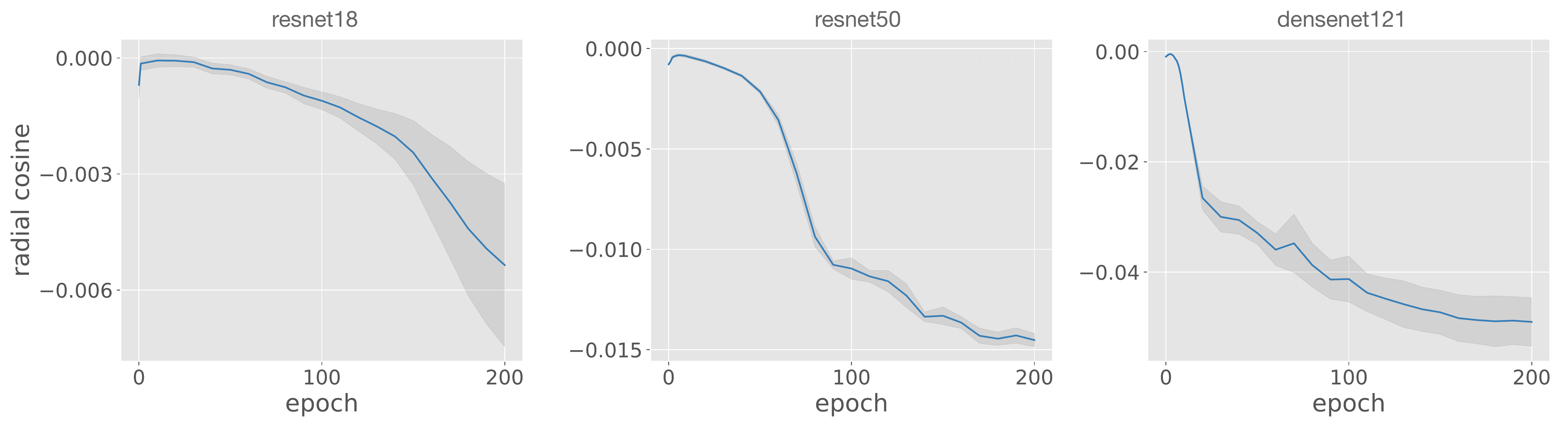}
\caption{The cosine between $w$ and $- \nabla \ell_{neg}$ for networks trained on shuffled labels. They are only negative, and the gradients w.r.t $\ell_{neg}$ do not contribute to norm growth along the radial direction.}
\end{figure}

\subsection{Experiments with Batch Normalization}

\begin{figure}[h]
\centering
\includegraphics[width=\textwidth]{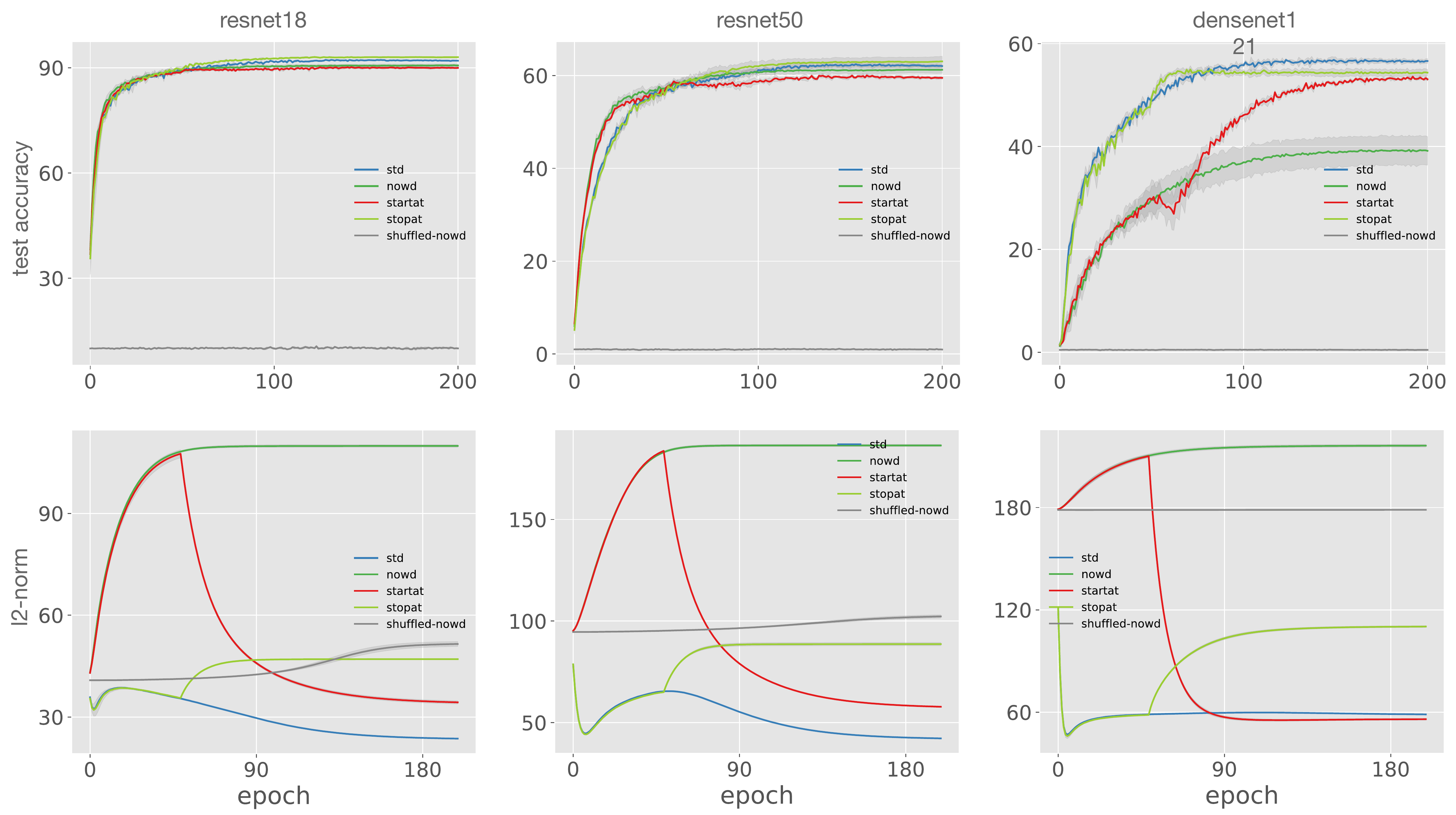}
\caption{The test accuracy and l2 norm of networks trained without batch normalization. Again, we see that using weight decay only at the start vastly outperforms using it after the start.}
\label{fig:app_nobn_acc}
\end{figure}

\begin{figure}[h]
\centering
\includegraphics[width=\textwidth]{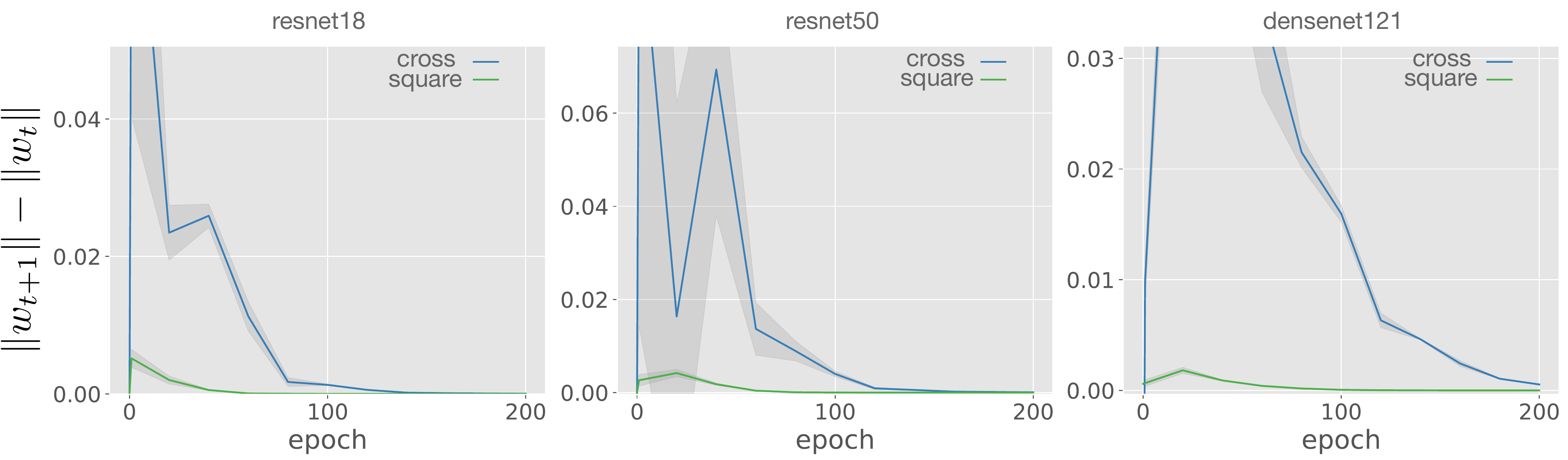}
\caption{The contributions of the cross and square term towards the norm growth for networks trained without batch normalization. The cross term dominates.}
\label{fig:app_nobn_cross_vs_sq}
\end{figure}

\begin{figure}[h]
\centering
\includegraphics[width=\textwidth]{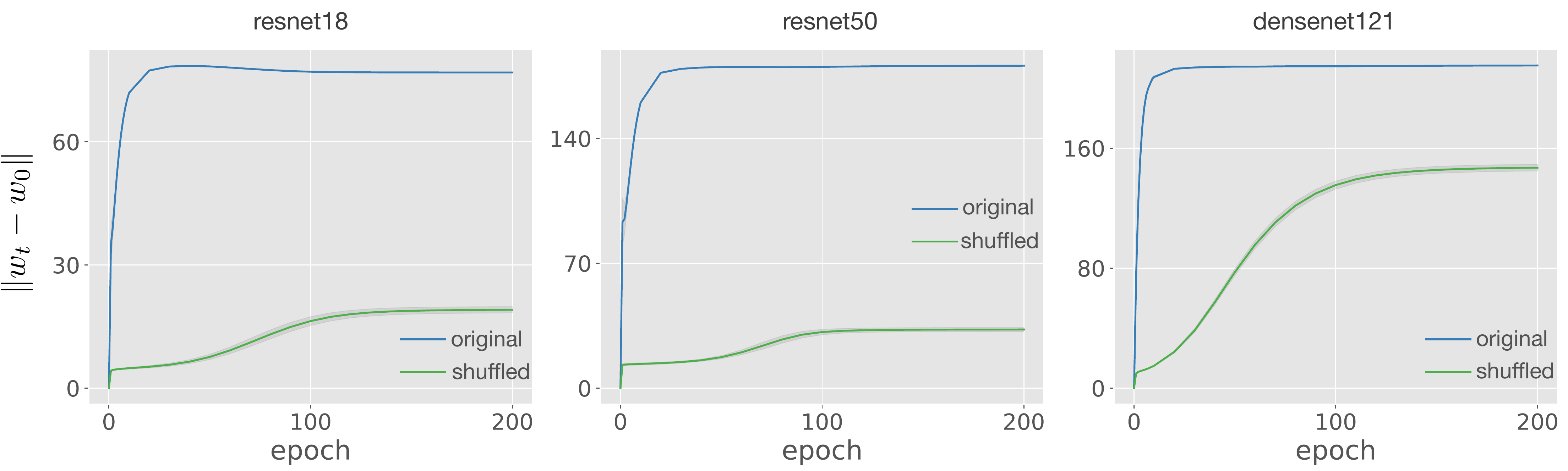}
\caption{The distance from start for networks without batch normalization trained on the original and shuffled labels. The networks move significantly.}
\label{fig:app_nobn_dist_from_start}
\end{figure}

\begin{figure}[h]
\centering
\includegraphics[width=\textwidth]{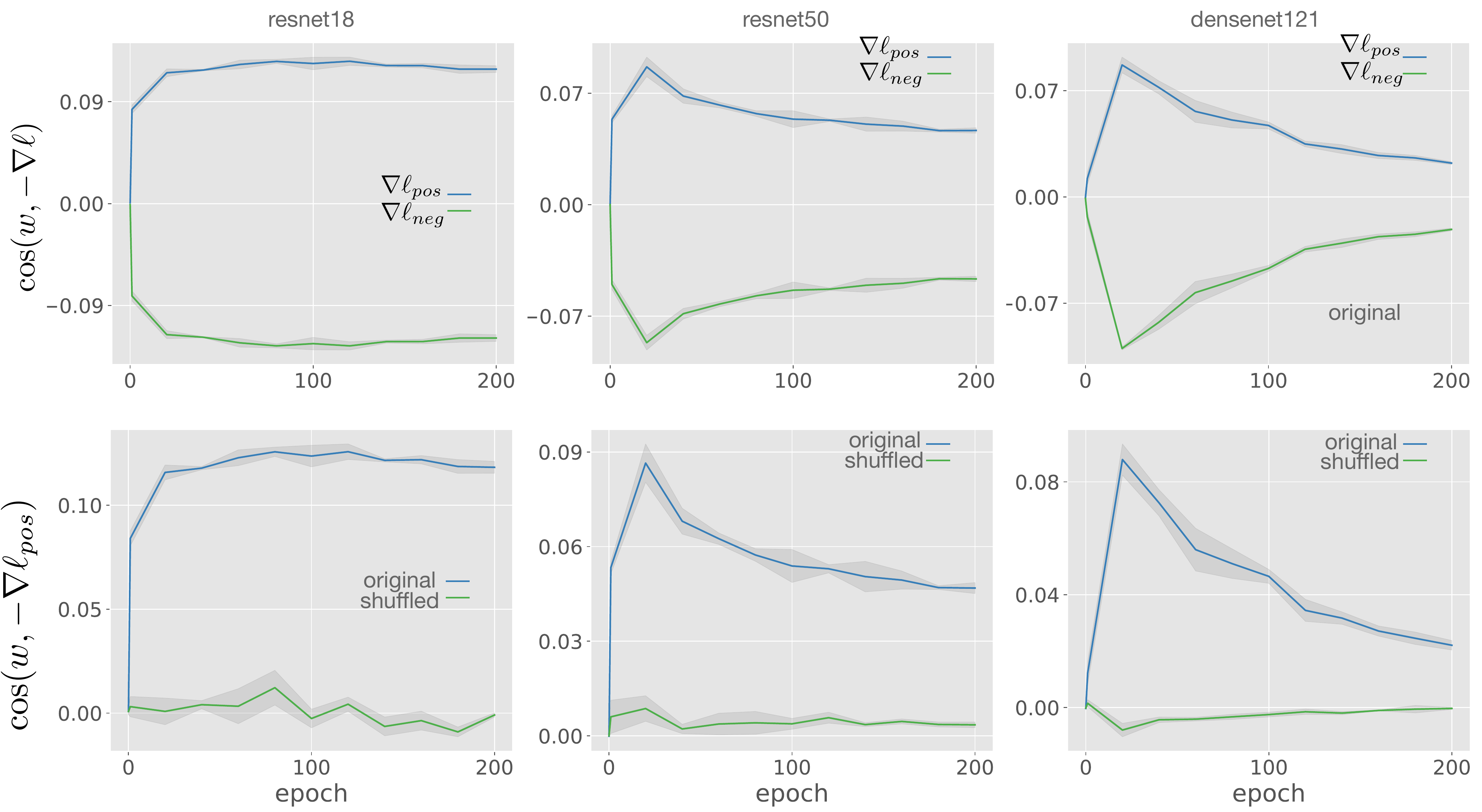}
\caption{Cosines between the weights and the gradients for networks without batch normalization. The results are similar to networks with batch normalization.}
\label{fig:app_nobn_grad_cosine}
\end{figure}

To investigate the relationship to batch normalization, we repeat experiments from the main paper with the same networks, but turn of batch normalization. These results are presented in \cref{fig:app_nobn_acc}, \Cref{fig:app_nobn_cross_vs_sq}, \Cref{fig:app_nobn_dist_from_start} and \Cref{fig:app_nobn_grad_cosine}.

\end{document}